\pdfoutput=1

\documentclass[11pt]{article}

\usepackage{acl}

\usepackage{times}
\usepackage{latexsym,amssymb}
\usepackage{booktabs}
\usepackage{amsmath}
\usepackage{comment}
\usepackage{graphicx}
\usepackage{stmaryrd}
\usepackage{bm}
\usepackage{dsfont}
\usepackage{xspace}
\usepackage{longtable}
\usepackage{inconsolata}
\usepackage[ruled]{algorithm2e}
\usepackage{multirow,multicol}
\usepackage{pdflscape}
\usepackage{tabularx}
\usepackage{afterpage}
\usepackage[capitalize]{cleveref}

\definecolor{darkgreen}{rgb}{0.0, 0.5, 0.0}
\definecolor{purple}{HTML}{7F3EFF}
\definecolor{blue}{HTML}{0076BA}
\definecolor{green}{HTML}{1DB100}
\definecolor{red}{HTML}{FF0000}

\usepackage[T1]{fontenc}

\usepackage[utf8]{inputenc}

\usepackage{microtype}

\usepackage{todonotes}

\title{Quantifying Adaptability in Pre-trained Language Models \\ with 500 Tasks}

\newcommand{\ourdataset}{\textsc{TaskBench500}\xspace}

\author{
    Belinda Z. Li \\ MIT \\ \texttt{bzl@mit.edu} \\
    \And Jane Yu \\ Meta AI \\ \texttt{janeyu@fb.com} \\
    \And Madian Khabsa \\ Meta AI \\ \texttt{mkhabsa@fb.com}
  \AND
      Luke Zettlemoyer \\ Meta AI \\ \texttt{lsz@fb.com} \\
      \And Alon Halevy \\ Meta AI \\ \texttt{ayh@fb.com} \\
      \And Jacob Andreas \\ MIT \\ \texttt{jda@mit.edu} \\
}

\begin{document}
\maketitle

\begin{figure*}
    \centering
    \includegraphics[scale=0.24,trim={0.5cm 12.5cm 0.5cm 0.4cm},clip]{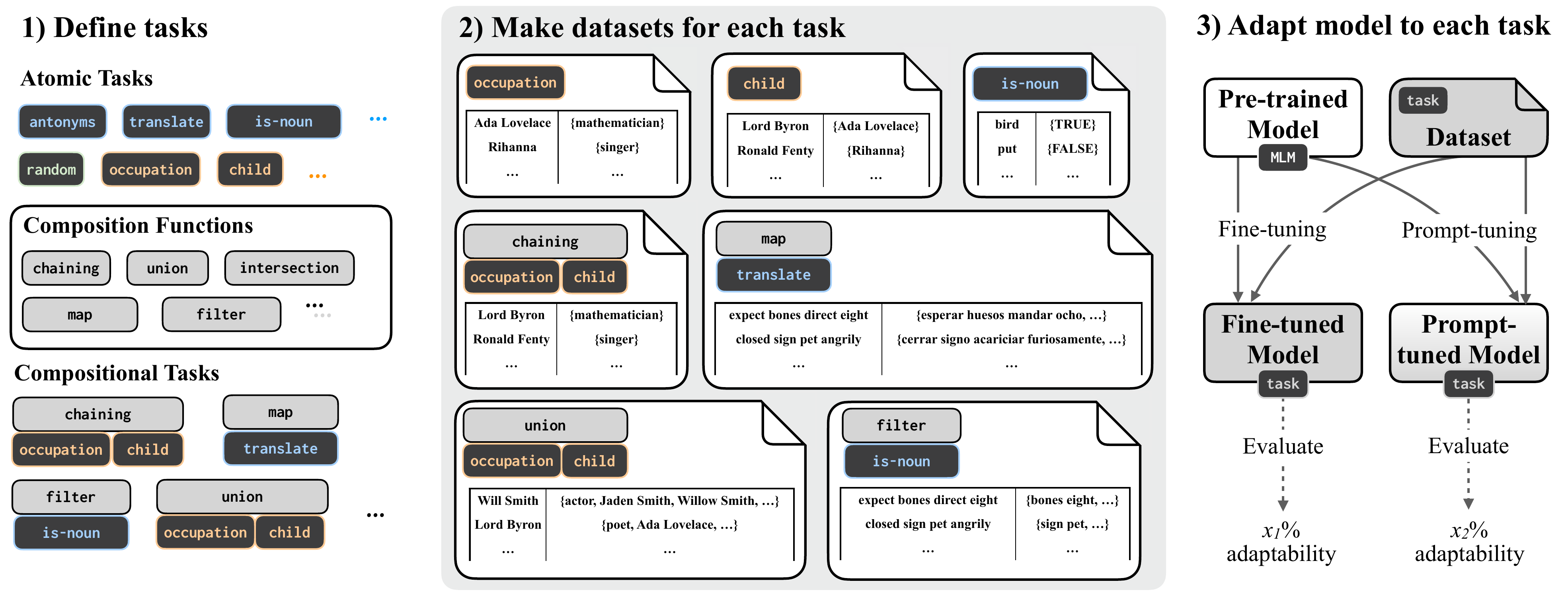}
    \caption{Overview of our task creation process. We begin by defining a set of atomic tasks that all synthetic tasks are built upon. These include lexical tasks (\textcolor{blue}{blue} text/outline), random tasks (\textcolor{green}{green} text/outline), and factual tasks (\textcolor{orange}{orange} text/outline). They also include both predicates and relations. These tasks are combined using composition functions to form more complex, compositional tasks. Given a particular task specification, we synthetically create a dataset for each task. Finally, we fine-tune or prompt-tune a pre-trained language model on each task dataset.}
    \label{fig:overview_figure}
\end{figure*}

\begin{abstract}
When a neural language model (LM) is adapted to perform a new task, what aspects of the task predict the eventual performance of the model? In NLP, systematic features of LM \emph{generalization} to individual examples are well characterized, but systematic aspects of LM \emph{adaptability} to new tasks are not nearly as well understood.
We present a large-scale empirical study of the features and limits of LM adaptability using a new benchmark, 
\ourdataset, built from 500 procedurally generated sequence modeling tasks. %
These tasks combine core aspects of language processing, including lexical semantics, sequence processing, memorization, logical reasoning, and world knowledge.
Using \ourdataset, we evaluate three facets of 
adaptability,
finding that: (1) adaptation procedures differ dramatically in their ability to memorize small datasets; (2) within a subset of task types, adaptation procedures exhibit \emph{compositional adaptability} to complex tasks; and (3) failure to match training label distributions is explained by mismatches in the intrinsic difficulty of predicting individual labels. Our experiments show that adaptability to new tasks, like generalization to new examples, can be systematically described and understood, and we conclude with a discussion of additional aspects of adaptability that could be studied using the new benchmark.

\end{abstract}

\section{Introduction}
Much of the recent research effort in NLP has shifted from training task-specific models to \emph{adapting} pre-trained language models (LMs) by fine-tuning their parameters or input prompts for downstream tasks~\cite{bert,T5,li2021prefix,lester2021power}. 
This paradigm is general, in the sense that a large number of distinct NLP tasks benefit from pre-training~\cite{ELMO,bert,T5}. But many questions about the nature and limits of LM adaptation remain unanswered. For example: given a new task, can we predict how quickly (and how effectively) pre-trained LMs can be adapted to perform it? From among the variety of different adaptation techniques (e.g.\ fine-tuning or prompt-tuning), can we predict which one will be most effective?
Today, new pre-training and adaptation schemes are evaluated using small suites of curated tasks, typically featuring %
classification, textual inference, and question answering ~\cite{wang2018glue,wang2019superglue}. These benchmarks have been extremely successful in identifying new tools for adaptation, but they are poorly suited for answering larger, structural questions like the ones above.

We present a large-scale study of LM adaptability using a new suite of benchmark tasks called \ourdataset.\footnote{Data and code available at: \url{https://github.com/facebookresearch/task_bench}}
\ourdataset consists of 500 \textbf{procedurally generated} tasks involving lexical semantics, factual information, memorization of random relations, list processing, and logical composition (\cref{fig:overview_figure}).
Analogous to past work that uses synthetic data to characterize LM performance on single examples~\cite{DBLP:journals/corr/WestonBCM15,lake2018generalization,DBLP:conf/iclr/SaxtonGHK19,kim-linzen-2020-cogs,keysers2020measuring,liu2021small}, \ourdataset enables systematic study of LM adaptability at the task level.
In this paper, we use it to study three aspects of adaptability:

\smallskip
\noindent \textbf{Memorization}: When can adaptation successfully memorize new functions (e.g.,\ to update factual knowledge about entities, or learn arbitrary new token correspondences)?
    We find that \textbf{LMs' ability to memorize new input--output mappings is strongly influenced by task type}. Datasets of lexical relations (like antonym pairs) are easier to memorize than factual information (like name--occupation pairs). Both are easier to memorize than lists of random word pairs.
    These findings are particularly striking in the case of prompt tuning, which in standard configurations struggles to memorize even small random word pair lists.
    
    \smallskip
\noindent
     \textbf{Composition}: Is LM performance on simple tasks predictive of their performance on compositions of those tasks? (If the \emph{father} and \emph{occupation} relations are easy to learn via adaptation, does this imply that the \emph{father's occupation} relation is also easy to learn?)
    We find a nuanced answer. \textbf{LMs exhibit compositional adaptation} to 
    lexical and factual relations (like \emph{father's occupation}), with success on composed tasks strongly correlated ($r^2 > 0.5$) with success on atomic tasks. 
    However, when composing these relations with sequence processing operations, success on the base task does not predict success on the composed task.
    
    \smallskip
\noindent
    \textbf{Distribution matching}: In models fine-tuned on datasets exhibiting a distribution of acceptable answers (e.g.,\ translating ungendered pronouns into gendered ones), do model predictions match these distributions? 
    We find that \textbf{LMs are often unable to 
    match label distributions in datasets used for adaptation}. In particular, when labels in the fine-tuning dataset are drawn from a uniform mixture of labels from two tasks (e.g.,\ labeling half of the words with their \textit{antonym} and half with their \textit{synonym}), models disproportionately assign mass to labels from the task that is easier to learn. %

Each of these forms of adaptability corresponds to a central challenge in NLP: reliable updating of deployed models, composition of previously learned skills, and fair and predictable output from models trained on curated data. 
Our study of memorization, composition, and distribution matching have direct analogs to previous studies of %
sample expressivity~\cite{DBLP:conf/iclr/ZhangBHRV17}, compositional generalization~\cite{lake2018generalization,kim-linzen-2020-cogs,keysers2020measuring}, and calibration~\cite{10.5555/3305381.3305518}.
However, we study these phenomena at the task level, rather than the example level.
Our experiments highlight important qualitative differences between current adaptation paradigms; identify several novel challenges for LM adaptation, and offer a new benchmark for approaches aimed at meeting these challenges.

\section{Background}

\paragraph{Fine-tuning and prompt search}
In languages for which large digitized corpora are available, most NLP system development today involves \emph{adaptation} of a pre-trained model to a downstream task of interest. Pre-training typically involves reconstruction of masked or corrupted text sampled from a large corpus~\cite{bert,liu2019roberta,T5}. Adaptation to a new task typically involves one of three approaches: (1)  \textbf{fine-tuning} of all of a pre-trained model's parameters (possibly in conjunction with a specialized decoder) on a task-specific training set~\cite{bert}; (2) manual \textbf{prompt engineering} of an input template that induces pre-trained model predictions to perform the task of interest~\cite{GPT3,petroni-etal-2019-language}; or (3) automated \textbf{prompt tuning} of these templates, in either the discrete space of tokens~\cite{autoprompt} or continuous space of token embeddings~\cite{li2021prefix,lester2021power,ptuningv2}. The latter two approaches have grown more popular as pre-trained models have grown larger.
The performance of both prompt-search approaches still lags fine-tuning~\cite{T5,GPT3,lester2021power}, though the difference between approaches appears to shrink as model size increases~\cite{lester2021power}.

\paragraph{Measuring generalization and adaptability}
The success of the training paradigm described above stems from its generality---a large number of NLP tasks appear to benefit from some combination of pre-training and adaptation. Previous attempts to \emph{quantify} this generality have typically relied on benchmarks like GLUE~\cite{wang2018glue} and SuperGLUE~\cite{wang2019superglue}, each of which aggregates ten natural language processing tasks designed to probe different aspects of language understanding. Similar benchmarks have also been built for non-English languages~\cite{xu-etal-2020-clue,kakwani-etal-2020-indicnlpsuite,park2021klue,DBLP:conf/icml/HuRSNFJ20}.
However, the heterogeneity and small number of distinct tasks represented in existing benchmarks makes it difficult to make finer-grained predictions, e.g.\ by identifying specific features of tasks that contribute to the success or failure of adaptation.

This challenge has a direct analog to the problem of characterizing \emph{generalization} at the example level in models trained for a single task. Model performance on natural test sets is often loosely correlated with accuracy on individual examples featuring rare syntactic constructions or word collocations~\cite{mccoy-etal-2019-right}. %
A great deal of past work has focused on improving evaluation using synthetic evaluation sets~\cite{DBLP:conf/emnlp/JiaL17,naik-etal-2018-stress,lake2018generalization,richardson2020probing}.
These datasets have been used to study long-range agreement~\cite{marvin-linzen-2018-targeted}, compositional generalization~\cite{lake2018generalization,NEURIPS2020_ruis,keysers2020measuring}, and mathematical reasoning~\cite{DBLP:conf/iclr/SaxtonGHK19}.
But no analogous notion of systematicity, or tool for studying it, currently exists for tasks rather than examples.

Thus, building on this past work, we describe how to
construct synthetic data distributions that enable systematic study of \emph{adaptation to new tasks} rather than \emph{generalization to new examples}. Like previous research that uses procedural data generation procedures to study models in NLP, we focus on coverage rather than naturalness, using datasets designed to complement, rather than replace, existing naturalistic benchmarks.

\section{A 500-task benchmark}

Our goal is to study the generalizability of task adaptation paradigms. In particular, we would like to identify which attributes of a task make it easy or difficult to learn, across different models and training schemes. 
While this work shares many of its high-level goals with existing benchmarks built from collections of real-world datasets, the makeup and difficulty of these datasets is often difficult to characterize precisely: differences in annotation standards, annotation quality, and dataset size mean that
models often exhibit very different performance on datasets designed to evaluate model performance on the same abstract task.
In addition, existing datasets cover an exceedingly small subset of the space of all tasks that future NLP practitioners might wish to perform.
To account for all these limitations, we propose to generate tasks \emph{synthetically} as described below.

\paragraph{The space of tasks} \ourdataset is constructed compositionally: we begin by defining a space of \textbf{atomic tasks},  which are combined using a set of \textbf{composition operators} to produce more complex tasks. 
Every task takes as input a word or word sequence, and outputs either a boolean value or a set of words/word sequences. 
We refer to any task that outputs booleans as a \textbf{predicate task}, and any task that outputs sets of words or word sequences as a \textbf{relation task}. A subset of relation tasks involve modeling relations between single words at the input and output; we refer to these as \textbf{word-level} tasks and the remaining relation tasks (that take sequences as input or output) as \textbf{sequential tasks}.

The choice of atomic tasks and composition functions aims to capture aspects of real language processing tasks.
Accordingly, the set of atomic tasks comprises of:
\begin{enumerate}
    \item \textbf{Lexical tasks}, which test knowledge of lexical semantics. These include \emph{lexical relations} like \texttt{synonym}, or \textit{lexical predicates} like \texttt{is-noun}. These tasks are constructed from WordNet relations~\cite{wordnet}.
    \item \textbf{Factual tasks}, which test factual knowledge. These include \emph{factual relations} like \texttt{father-of}, or \emph{factual predicates} like \texttt{is-human}. These tasks are constructed from Wikidata properties~\cite{wikidata}.  
    \item \textbf{Random relation tasks}, which test memorization ability. These are created by mapping a word in the vocabulary to a singleton set containing a random other word. 
    We create 4 random relations with different random seeds.
\end{enumerate}
To recursively create arbitrarily complex tasks, we define a set of \textit{composition functions}. These take tasks as arguments and return other tasks. These functions fall into two categories: %
\begin{enumerate}
    \item \textbf{Word-level compositions}, which test ability to combine word-level information in different ways, such as through set or logical operations. 
    These functions take word-level tasks and return other word-level tasks.
    Examples include \texttt{intersection} and \texttt{chaining}. 
    \item \textbf{Sequential compositions}, which test ability to operate on sequences.
    These functions convert word-level tasks to sequence-level tasks. There are two functions in this category: \texttt{map} takes a word-level relation task 
    and returns a task that maps a sequence of $n$ words to a set of all possible sequences resulting from applying $f_W$ to each input word.\footnote{Note word-level relations return \emph{sets} of words---we turn a sequence of sets of words into a set of sequences by considering all combinations of words in each set.}
    \texttt{filter} takes word-level predicate tasks 
    and returns a sequence consisting only of words for which the task returns $\mathtt{true}$, preserving the original ordering of those words.
\end{enumerate}
The full list of atomic tasks and composition function can be found in Appendix~\cref{tab:atomic_tasks,,tab:comp_functions}.
We surmise that typical NLP tasks may require some combination of %
lexical knowledge, factual knowledge, sequential processing, and other task-specific reasoning;
our data distribution lets us evaluate all these aspects separately and in combination.

\paragraph{Datasets for tasks} 
We create datasets $\mathcal{D}(f) = \{(x_i, y_i) : x\sim\mathcal{X}_f, y \sim \textrm{Unif}(f(x_i)) \}$ for each task $f$, where $\mathcal{X}_f$ is the input distribution for the task, and recalling that $f(x_i)$ returns a \emph{set} of possible outputs associated with the input $x_i$.
For all tasks, we randomly partition the dataset into $\mathcal{D}_\text{train}(f)$ and $\mathcal{D}_\text{test}(f)$ splits.

For \textit{lexical atomic tasks} and their compositions, we directly use the most common words in the task's input language for $\mathcal{X}_f$. We create tasks in English and Spanish. %
For \textit{factual atomic tasks} and their compositions, we sample the entities from Wikidata that participate in the relation or predicate defined by the task (e.g.\ for the \texttt{child-of} task, we sample only entities that have children).
For \textit{sequential tasks}, we use a random sampler, which samples $n$ random words from the vocabulary and concatenates them.

\Cref{fig:overview_figure} shows examples of tasks and associated datasets.
More details on dataset construction can be found in~\cref{sec:appendix_dataset_details}.

\begin{figure*}
    \centering
    \includegraphics[scale=0.3,trim={0.5cm 21.4cm 16cm 0},clip]{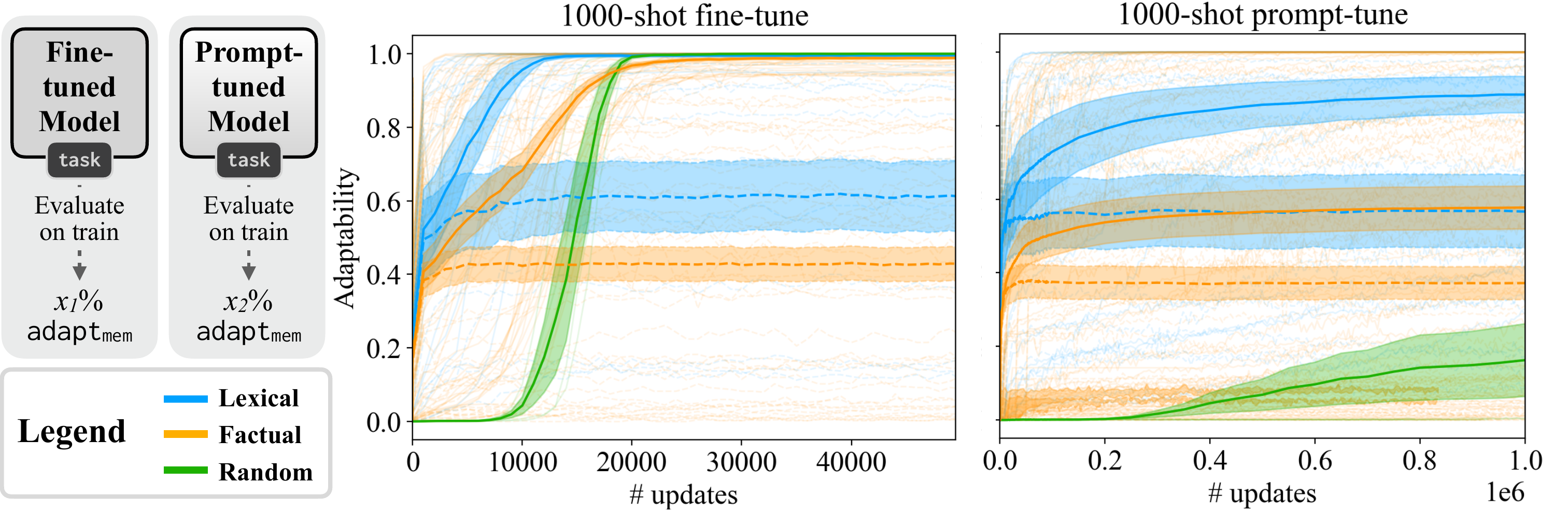}
    \caption{Left: Overview of the memorization experiment, which evaluates how accurately models adapted via fine-tuning and prompt-tuning can memorize training data.
    Right: Memorization and generalization curves for fine-tuning and prompt-tuning on 1000 training examples. Memorization curves are shown by solid lines, while generalization curves are dashed.
    We average over all atomic tasks from each task category: \textcolor{blue}{lexical} tasks, \textcolor{orange}{factual} tasks, and \textcolor{green}{random} tasks.
    The shaded region shows the standard error of the mean. Transparent lines are each individual task, colored by task category.
    In both paradigms, lexical tasks are easiest to memorize, followed by factual tasks, then random tasks. However, prompt-tuning has overall much less memorization capacity than fine-tuning, which can perfectly memorize even completely random relations.
    }
    \label{fig:memorization}
\end{figure*}

\section{Experimental Setup}
\paragraph{Model \& Training}
For all experiments, we adapt a pre-trained T5-base model~\cite{T5}. %
We examine two types of training paradigms: fine-tuning and prompt-tuning. 
During fine-tuning, we update all model parameters on the training set. 
During prompt-tuning, we follow~\newcite{lester2021power} and introduce a new set of prompt-tokens $\{p_1,\cdots,p_n\}$ to the vocabulary, which will be prepended to every sample from the task during inference, i.e., each sample input $x$ becomes $p_1p_2\cdots p_nx$. 
Let $\bm{\theta}$ denote the parameters of the original pretrained LM.
During training, the entire model is frozen and only the word embeddings of the new tokens $\{\theta_{p_1},\cdots, \theta_{p_n}\}\subset \bm{\theta}$ are updated. 
We use prompts of length $n=100$ for all experiments. We also study each paradigm on various quantities of training data, %
and separately evaluate their memorization and generalization adaptabilities. In particular, for word-level tasks the test-set words are disjoint from the train-set words, so evaluating on the test set will strictly measure generalization capacity.
We optimize all models using AdamW. See~\cref{sec:appendix_training_details} for full hyperparameters.

\paragraph{Evaluation}
For each task $f$ and model $\mathcal{M}[\bm{\theta}]$ (with parameters $\bm{\theta}$), we measure the model's average per-token accuracy on both training and test splits of the dataset $\mathcal{D}(f)$.
As each task defines multiple acceptable outputs for each input, we credit models for producing any acceptable output.
Letting $y' = \mathcal{M}(x)$,
we measure the fraction of positions $i$ at which any valid answer $y_i$ matches the predicted $y'_i$:
\begin{align}
    &\mathtt{acc}(\mathcal{M}, \mathcal{D}(f)) =
    \nonumber \\ 
    &\qquad \max_{y\in f(x)} \frac{1}{|\mathcal{D}|} \sum_{(x,y)\in \mathcal{D}} \left( \frac1n \sum_{i=1}^n \llbracket y_i = y'_i \rrbracket \right)
    \label{eq:accuracy}
\end{align}
Further details can be found in~\cref{sec:segmentation_seq}.

Given a pretrained model $\mathcal{M}[{\bm{\theta}_{\mathtt{pretrain}}}]$, an adaptation procedure $\mathcal{T}$, and a task suite $f$, let $\mathcal{M}[{\bm{\theta}_{\mathcal{T},\mathcal{D}(f)}}]$ denote the model trained using $\mathcal{T}$ on training data $\mathcal{D}(f)$.
We then define the \textit{adaptability} of a (pretrained model, adaptation paradigm, task suite) as:
\begin{align}
    \texttt{adapt}&(
    \mathcal{M}[\bm{\theta}_{\texttt{pretrain}}], \mathcal{T}, f) \nonumber \\
    &= \texttt{acc}(\mathcal{M}[{\bm{\theta}_{\mathcal{T},\mathcal{D}_\text{train}(f)}}], \mathcal{D}_\text{eval}(f))
    \label{eq:adaptability}
\end{align}
We denote by $\texttt{adapt}_\text{mem}$ the value of this metric over training data ($\mathcal{D}_\text{eval} = \mathcal{D}_\text{train}$), and by $\texttt{adapt}_\text{gen}$ the metric over test data ($\mathcal{D}_\text{eval} = \mathcal{D}_\text{test}$).

\section{Memorizing datasets}
\label{sec:memorization}

Our first experiment investigates the extent to which task adaptation paradigms can memorize different types of tasks. We are interested in memorization because many  real NLP tasks involve some degree of memorization. For example, translation builds on memorizing lexical associations between words in various languages, and semantic similarity and paraphrasing require memorizing word meanings and/or groupings of semantically similar words.

\paragraph{Method}
We use training-set adaptability ($\texttt{adapt}_\text{mem}$) as an indicator of a model's memorization ability (\cref{fig:memorization}).
We train on a set of 1000 examples, and plot the value of~\cref{eq:adaptability} on each \textit{atomic} task as models are adapted via fine-tuning or prompt-tuning.
This allows us to visualize both the final training-set performance, as well as the time it took to achieve that performance, both of which we use to quantify memorization ability.

\paragraph{Results}
\Cref{fig:memorization} shows the training curves for fine-tuning (left) and prompt-tuning (right), on different types of tasks. Solid lines show $\texttt{adapt}_\text{mem}$, while dashed lines show $\texttt{adapt}_\text{gen}$.

Under both adaptation paradigms, we find that lexical tasks are easier to memorize than factual tasks, while random tasks are the hardest to memorize.
However, for fine-tuning, we find that models can (eventually) learn to perfectly memorize all types of tasks---even entirely random word associations. 
However, different types of tasks converge at different rates---lexical tasks converge first, followed by factual tasks, followed by random tasks.

Prompt-tuning, with many fewer parameters than fine-tuning, is much less expressive. 
As shown in~\cref{fig:memorization}, none of the tasks types converge to 100\% accuracy across tasks. Prompt-tuning overall also takes significantly longer to converge; in particular, on random tasks, the finetuned model generally converges at $\sim$20k updates, while the prompt-tuned model takes over 200k updates to even begin performing nontrivially.

However, despite being much worse at memorization, prompt-tuned models still generalize almost as well as fully fine-tuned models, at least on atomic tasks. 
This suggests that the inability to memorize arbitrary functions is not %
necessarily a problem for prompt-tuning in general,
and more broadly that overfitting the training set---at least during fine-tuning---may not be necessary for generalization.

In~\cref{sec:app_permute}, we run a version of this experiment on \textit{permuted} task labels in order to better disentangle the effect of learning novel tasks vs. retrieving existing ones. We find that, for both prompt-tuning and fine-tuning, pre-trained models can more easily adapt to existing relations than to novel (permuted) ones, but they are still \textit{able} to adapt to new tasks, especially compared to non-pre-trained models.

\begin{figure*}
    \centering
    \includegraphics[scale=0.3,trim={0 13.7cm 14cm 0}, clip]{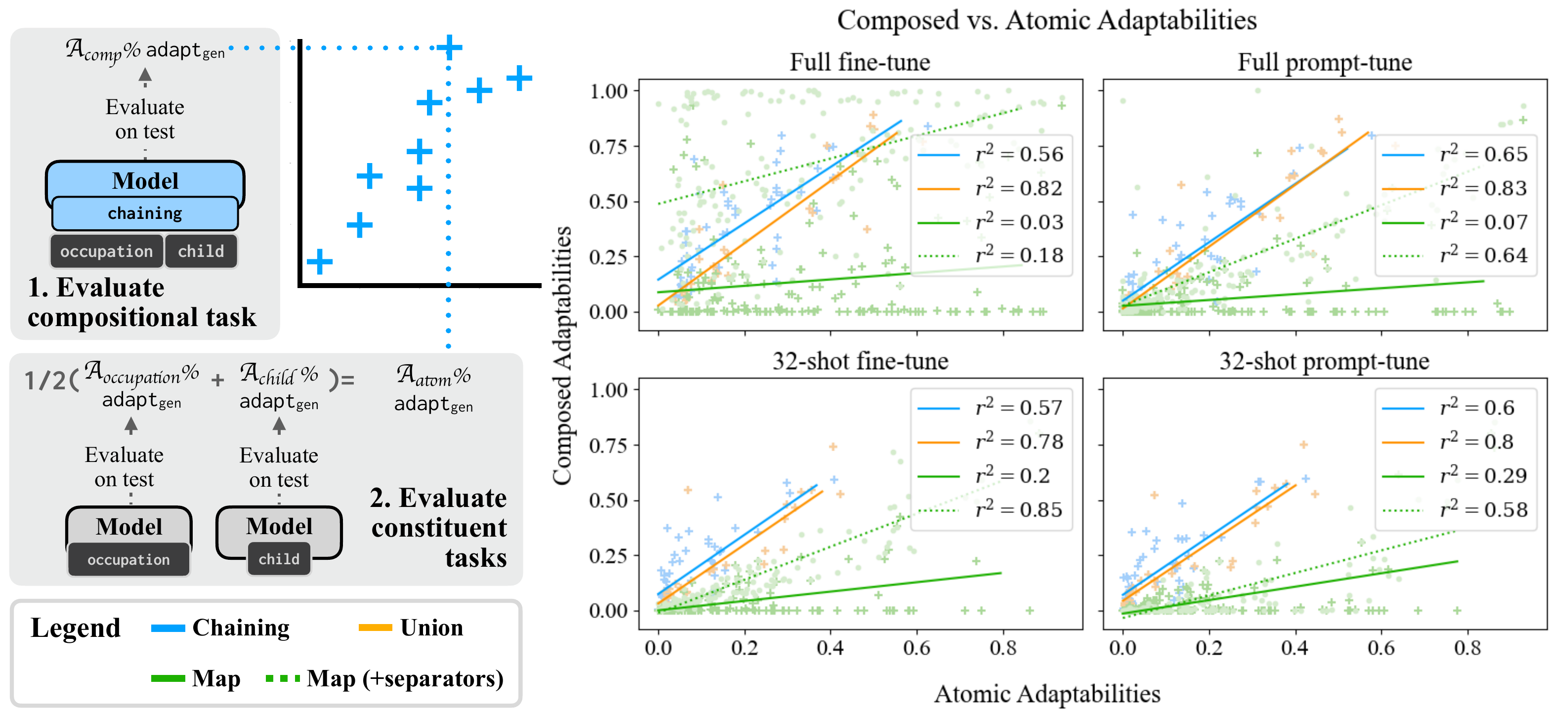}
    \caption{
    Left: Overview of the composition experiment. We evaluate how well the adaptability on a compositional task can be predicted by the (averaged) adaptabilities of the atomic constituent tasks. Right: 
    Correlation between compositional adaptability vs. averaged atomic adaptabilities, for the \textcolor{blue}{chaining}, \textcolor{orange}{union}, and \textcolor{green}{map} composition types, under each training paradigm.
    On word-level \textcolor{blue}{chaining} and \textcolor{orange}{union} compositions, compositional adaptability is observed: composed task performance is highly correlated with atomic task performance ($r^2 > 0.5$) under all training paradigms.
    However, on sequential \textcolor{green}{map} compositions, all models perform poorly, and thus non-compositionally.
    This results from challenges in segmenting input sequences; if token boundaries are explicitly marked
    (\textcolor{green}{map (+separators)}),
    compositional adaptability is again observed.
    }
    \label{fig:composition}
\end{figure*}
\begin{table}[]
    \centering
    \small
    \begin{tabular}{cccc}
    \toprule
        & Atomic & Word-level Comp & Seq Comp \\
    \midrule
        FFT & $46.9_{\pm 4.0}$ & $39.5_{\pm 2.1}$ & $21.5_{\pm 1.9}$ \\
        FPT & $42.6_{\pm 4.3}$ & $28.1_{\pm 2.4}$ & $11.5_{\pm 1.4}$ \\
        32FT & $33.6_{\pm 3.8}$ & $22.2_{\pm 1.8}$ & $5.7_{\pm 0.9}$ \\
        32PT & $32.4_{\pm 3.6}$ & $21.7_{\pm 1.7}$ & $6.9_{\pm 1.1}$ \\
    \bottomrule
    \end{tabular}
    \caption{Model (generalization) adaptabilities to atomic, word-level compositional, and sequential compositional tasks, under full fine-tuning (FFT), full prompt-tuning (FPT), 32-shot fine-tuning (32FT) and 32-shot prompt-tuning (32PT).
    Prompt-tuned models are comparable to fine-tuned models for atomic tasks, but not for compositional tasks. However, this distinction disappears under few-shot learning.
    }
    \label{tab:comp_tasks_acc}
\end{table}

\section{Composing tasks}
\label{sec:compositionality}

In the previous section, we found that while prompt-tuning cannot memorize arbitrary tasks like fine-tuning, it can still generalize well on simple atomic tasks, almost comparably to fine-tuning. In this section we investigate whether this finding extends to more complex tasks. Specifically, we examine the behavior of prompt-tuned and fine-tuned models when adapted to \textit{compositions} of atomic tasks.

Many prior studies of compositionality focus on \textit{instance-level} compositionality~\cite{lake2018generalization,keysers2020measuring}: they test whether models can generalize to new instances by combining information from previously-seen instances \textit{within the same task}. For example,~\newcite{lake2018generalization} study whether models can learn to \emph{jump left}, after learning to \emph{jump}, \emph{run}, and \emph{run left}. In our work, we instead focus on \textit{task-level} compositionality, studying whether models can adapt to new \textit{tasks} that are compositions of simpler tasks on which they are known to perform well.
Thus, while a model exhibiting \textit{compositional generalization} will correctly compose fragments of previously observed training examples, a training procedure exhibiting \textit{compositional adaptability} will perform well on tasks involving compositions of previously learned skills.

\paragraph{Method} %
We study adaptation to complex tasks by 
relating performance on \emph{atomic} tasks with performance on
\textit{depth-2 compositional tasks}. 
We also study each paradigm under few-shot learning, by creating a random 32-sample subset of each training dataset, and training on that subset. %
To mitigate the effect of the random seed, we report average performance over 4 different subsets. %

What allows models to adapt to these complex tasks? We hypothesize that their adaptability is (in part) compositional---when they can adapt to simple tasks, they can also adapt to compositions of those tasks.
For each training paradigm $\mathcal{T}$ and each composition function $C$, we 
run linear regression to
estimate the Pearson correlation coefficient $r^2$ between adaptability to a compositional task $C(f_{1}, \cdots f_{n})$,
\begin{align}
\mathtt{adapt}_\text{gen}(\mathcal{M}, \mathcal{T}, C(f_{1}, \cdots, f_{n})),
\end{align}
and \textit{average} adaptability to the task's atomic components,
\begin{align}
\frac1n \sum_{i=1}^n \mathtt{adapt}_\text{gen}(\mathcal{M}, \mathcal{T}, f_{i}).
\end{align}
\Cref{fig:composition} depicts the procedure graphically.\footnote{We focus only on compositional functions $C$ which have at least 20 compositional tasks $C(f_1,\cdots, f_n)$ in \ourdataset, so that we have at least 20 points to obtain a statistically significant correlation coefficient.}

\paragraph{Can language models learn compositional tasks?}
The average model adaptability to compositional and atomic tasks, under each training paradigm, is reported in~\cref{tab:comp_tasks_acc}. We observe that the gap between full-data prompt-tuned models and full-data fine-tuned ones is much larger on compositional tasks than atomic ones. %
Thus, prompt-tuned models can only generalize comparably to finetuned ones for sufficiently ``simple'' tasks. %

Interestingly, this distinction disappears under few-shot learning. Though both adaptation paradigms generalize much worse in the few-shot setting compared to the full setting, they appear to be comparable to each other in the few-shot setting, even on compositional tasks.
This may simply imply that few examples are insufficient to learn the nuances of complex tasks, and that simply learning a few prompt tokens is sufficient to capture what can be learned from the limited data samples.

\paragraph{Do language models adapt compositionally?}
We visualize each regression model in~\cref{fig:composition}. %
Higher $r^2$ indicates higher correlation between atomic and compositional versions of tasks.
Note that \textit{all} model training paradigms demonstrate some degree of word-level compositionality ($r^2 > 0.5$)---when they succeed at word-level compositional tasks (\texttt{union}, \texttt{chaining}), they succeed at the atomic constituents to those tasks, and vice versa.
However, this does not appear to be the case for sequential \texttt{map}. In the full-data regime, both fine-tuning and prompt-tuning have near-zero $r^2$ values. 
In the few-shot regime, the $r^2$ value, while nontrivial, is also quite low.
Note the slopes of the learned regression lines---the model appears to be unable to learn the sequential versions of tasks, despite succeeding at their atomic versions.

To explain this result, we hypothesize that a major obstacle to sequence-level compositional adaptability is \emph{segmentation} of sequences into atomic units. This is especially the case for factual tasks: for example, the sequence \texttt{Pauline Payne Whitney Charles Lloyd} could be segmented as \texttt{[Pauline Payne Whitney] [Charles Lloyd]} or \texttt{[Pauline Payne] [Whitney Charles Lloyd]}, etc.
To test whether segmentation is a bottleneck, we train on a version of sequential tasks where we give the language model \textit{explicit} markers of word/entity boundaries (e.g. the language model is given \texttt{Pauline Payne Whitney \# Charles Lloyd} as input).
We found that, with separators, performance on the \texttt{map} tasks increases substantially and the model demonstrates compositional adaptability ($r^2 > 0.5$) to these tasks in 3 of the 4 adaptation paradigms. This setting is plotted in~\cref{fig:composition} as \emph{Map (+separators)}.

\begin{figure*}
    \centering
    \includegraphics[scale=0.3,trim={0.5cm 17cm 21cm 0.5cm},clip]{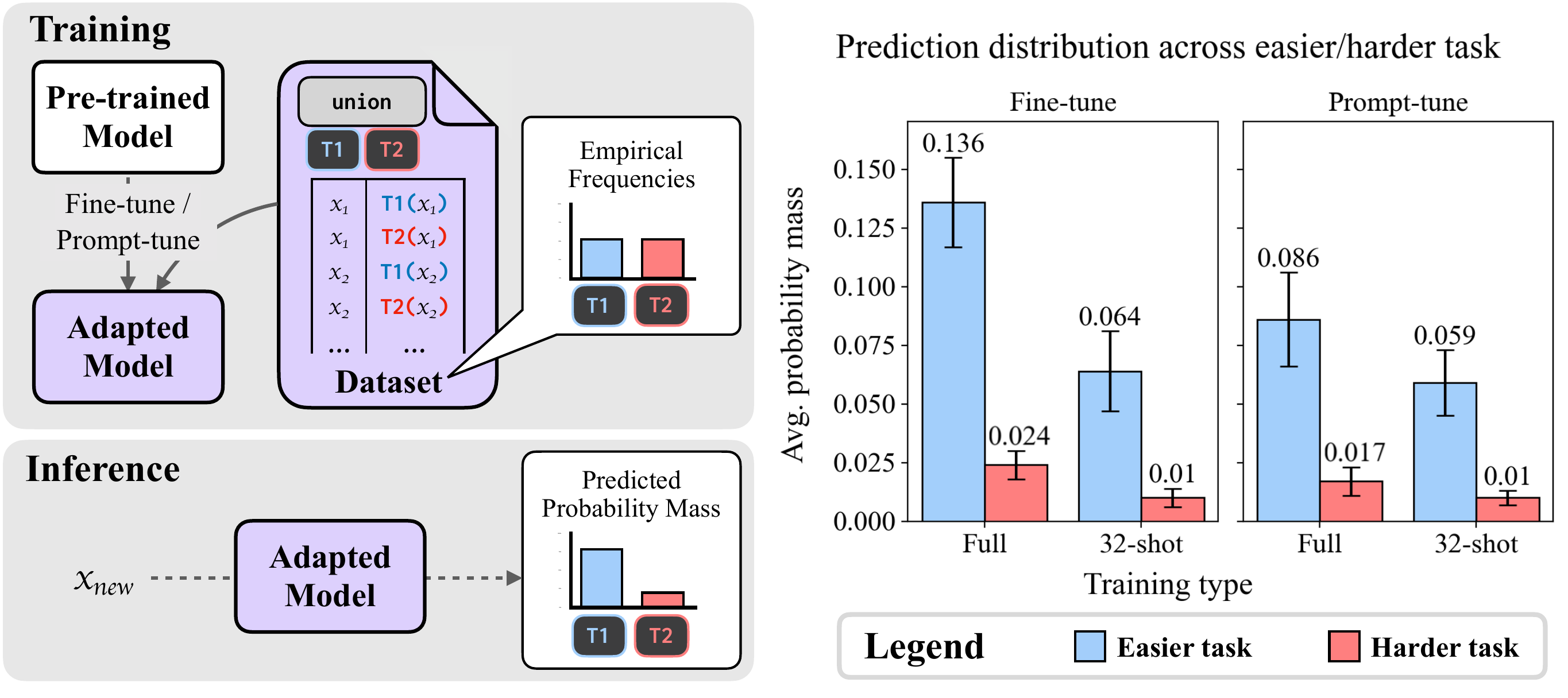}
    \caption{Left: Overview of the task prediction distribution experiments (\cref{sec:distribution_experiments}). We train a model on a balanced dataset of two tasks, and check whether the prediction distribution over tasks on novel examples matches the (balanced) empirical distribution.
    Right: Probability mass, across all pairs of tasks, assigned to all answers corresponding to the \textcolor{blue}{easier} vs. \textcolor{red}{harder} task, when trained on a balanced dataset and evaluated novel examples.  We report the average across all task pairs and held-out examples, as well as standard errors for each task pair. Note that the model tends to assign more probability to the \textcolor{blue}{easier} task, despite the task training set being balanced.
    }
    \label{fig:distribution}
\end{figure*}
Under this setting, full fine-tuning is the only training paradigm that does not demonstrate compositional adaptability. %
To better understand this phenomenon, we exclusively plot points from the \textit{Map (+separators)} setting in Appendix~\cref{fig:composition_septokens}. %
We find that the distribution of points in the full fine-tuning case shows that points tend to fall within the upper-left triangle.
This indicates that for a significant number of tasks, \textit{models adapt to their sequential versions despite failing at atomic versions.}
In these cases, the model does not simply adapt compositionally, but can take advantage of additional information present in sequences (e.g., seeing more tokens, more examples of the word-level function) to outperform compositional adaptation.

\section{Learning new distributions}
\label{sec:distribution_experiments}
Previous sections investigated the degree to which models could fit particular tasks
using a binary metric that assigned credit to any acceptable answer.
Our final set of experiments explores a finer-grained notion of correctness: when there are multiple acceptable answers, as is often the case in real NLP tasks, when does the output 
\textit{distribution} of a model match the distribution empirically observed during adaptation?

\paragraph{Method}

We specifically investigate whether models are biased towards predicting ``easy'' labels, in the sense measured in~\cref{sec:memorization}.
We consider all possible pairs of atomic tasks $f_1, f_2$ (for which $f_1$ and $f_2$ take in overlapping sets of inputs). %
Let $f_e$ to be the easier task in this pair
and $f_h$ be the harder task, relative to a model $\mathcal{M}$ and training paradigm $\mathcal{T}$, in the sense that $\mathtt{adapt}_\text{gen}(\mathcal{M}, \mathcal{T}, f_e) > \mathtt{adapt}_\text{gen}(\mathcal{M}, \mathcal{T}, f_h)$.
We compose $f_e$ and $f_h$ using \textit{union} to create compositional task $\cup (f_e, f_h)$, and construct the training dataset for this task to be balanced --- such that the model sees an equal number of examples of form $(x, f_e(x))$ as $(x, f_h(x))$.
Now let $\mathcal{M}_{\cup(f_e,f_h)}$ denote a model adapted to this task.
During test-time, we provide $\mathcal{M}_{\cup(f_e,f_h)}$ with novel inputs $x'$ from the domain of both $f_e$ and $f_h$, and 
record the average probability mass it assigns to all $y_e^i\in f_e(x')$ %
versus all $y_h^i\in f_h(x')$%
.\footnote{Note that the model may (and often does) assign mass to answers outside of these sets.}

Finally, we average %
these dataset-wide probabilities
over all pairs of tasks, to get an aggregated probability mass assigned to all easier tasks and all harder tasks in a task pair, invariant of the actual underlying task identity. %
More details on this procedure can be found in~\cref{sec:app_union}.

\paragraph{Results}
Overall, as seen in~\cref{fig:distribution}, across all tasks and training paradigms, the model tends to assign a higher probability to the easier relation. 
As a concrete example, when trained to predict either antonyms or lexical entailments, the average probability mass placed on the antonyms of a word from the held-out set (easier relation) is 13\%, while the average probability mass placed on the entailments of a word (harder relation) is 8\%.

Thus, despite having a perfectly balanced fine-tuning set, pretrained models still predict label distributions in a way that align with their inductive biases (measured via the ``intrinsic difficulty'' of individual labels). %
This holds for all task adaptation methods, including full fine-tuning,
meaning even paradigms and models that \textit{can} fit more complex tasks still have residual biases from pretraining that affect their predictions. 
This also suggests wider-reaching consequences for model fairness and equity: simply debiasing a fine-tuning dataset is insufficient to overcome biases from pretraining.

\section{Conclusion}
In this paper, we construct \ourdataset, a synthetic task set which serves as a testbed for task adaptability.
We focus on three axes of adaptability: ability to memorize, ability to (compositionally) generalize, and ability to fit to novel distributions.
We study two adaptation paradigms: fine-tuning and prompt-tuning, finding that: 1. unlike fine-tuning, prompt-tuning 
cannot memorize completely arbitrary tasks beyond a small number of examples,
2. all adaptation paradigms demonstrate compositional adaptation to word-level compositions, %
but not sequence-level compositions,
and 3. no paradigm is able to perfectly replicate the downstream distribution---all paradigms learn output distributions that align with its inductive biases.

In future work,
\ourdataset can be used to study other factors that may affect adaptability, such as length of the prompt in prompt-tuning, similarity between the task distribution and the pre-training distribution, %
or finer-grained distinctions between tasks (beyond lexical/factual/random, or composition type) that predict task adaptability.
\ourdataset can also be used to
explore the limitations of \textit{prompt engineering} on a GPT3-scale model. %
Finally, the current set of tasks and primitives in \ourdataset 
are by no means complete. Future work can expand on these primitives and study the relationships between the tasks put forth here and real NLP tasks.

\section*{Acknowledgements}

This work was supported by the MIT–IBM Watson AI lab. Part of the work was done using computing resources provided by a hardware donation from NVIDIA under the NVAIL program, and by the Lincoln Laboratory Supercloud. BZL is supported by an NDSEG Fellowship.

\section*{Impact Statement}
This paper introduces a new procedure for defining task suites. This procedure is then used to create a 500-task benchmark, which measures the adaptability of pre-trained language models to new tasks.
Because the benchmark is created procedurally from databases of words and entities, we anticipate that there should be little to no identifying information or toxic and hateful content. Our datasets should also contain less social bias compared to natural datasets.

However, like with all benchmarks, overfitting to static datasets can inhibit progress in NLP. 
Moreover, even though this dataset is procedurally generated, we cannot avoid all biases. The resources upon we build our benchmark are themselves biased---for example, lexical databases (like WordNet) are much richer for certain languages (like English) than others, and WikiData currently features many more men than women.
Our benchmark currently only features English and Spanish tasks, with a heavy bias towards standard English. This can encourage development of methods that underserve non-standard-English-speaking communities.

We hope to mitigate the aforementioned issues by releasing the code to procedurally generate task suites. We emphasize that the benchmark is dynamic: consisting of not just the static task suite that we are currently releasing, but more importantly the procedure for creating new tasks suites.
We encourage future researchers to develop analogous task suites for low-resource languages, non-standard English dialects, and more balanced sets of entities.

\bibliography{anthology,custom}
\bibliographystyle{acl_natbib}

\newpage
\appendix
\label{sec:appendix}

\begin{table*}[]
    \centering
    \small
    \begin{tabular}{cc}
    \toprule
        Task ($T$) & SPARQL fragment (\textcolor{blue}{\texttt{sparql(}$T$\texttt{, y)}}) \\
    \midrule
        \texttt{A(x)} & \texttt{?x A ?y .} \\
        \texttt{union(T1(x),T2(x))} & \texttt{\{ \textcolor{blue}{sparql(T1(x), y)} \} UNION \{ \textcolor{blue}{sparql(T2(x), y)} \}} \\
        \texttt{intersection(T1(x),T2(x))} & \texttt{\textcolor{blue}{sparql(T1(x), y) sparql(T2(x), y)}} \\
        \texttt{lor(T1(x),T2(x))} & \texttt{BIND( y1 || y2 AS y ) \textcolor{blue}{sparql(T1(x), y1)} \textcolor{blue}{sparql(T2(x), y2)}} \\
        \texttt{land(T1(x),T2(x))} & \texttt{BIND( y1 \&\& y2 AS y ) \textcolor{blue}{sparql(T1(x), y1)} \textcolor{blue}{sparql(T2(x), y2)}} \\
    \bottomrule
    \end{tabular}
    \caption{Rules for mapping word-level factual tasks to SPARQL conditional statements. \textcolor{blue}{Blue} substrings represent recursive calls to this set of rules, which are to be replaced with their output SPARQL fragments. Note the second argument to the \texttt{sparql} function represents the variable name to output to.}
    \label{tab:sparql_formula}
\end{table*}

\section{More details on \ourdataset creation procedure}
\label{sec:appendix_dataset_details}
\subsection{Task creation details}
For atomic lexical tasks, we take a subset of relations specified in either Wordnet~\cite{wordnet} or SentiWordNet~\cite{sentiwordnet}. For atomic factual tasks, we take a subset of tasks from Wikidata~\cite{wikidata}. We also have 3 broad categories of composition functions: set operations, logical operations, and sequential operations.
The full list of atomic tasks can be found in~\cref{tab:atomic_tasks} and the list of composition functions can be found in~\cref{tab:comp_functions}.

We enumerate all possible depth-2 word level compositions of each task, and the sequential versions of them (i.e. if the task is a relation, inserting it into a \texttt{map}, or if the task is a predicate, inserting it into a \texttt{filter}), up to 500 tasks. We also apply some basic heuristics to filter identical tasks: for example, we filter symmetric relations, e.g. \texttt{union(B,A)} is identical to \texttt{union(A,B)}, or avoid the use of logical operations alongside set operations, e.g. \texttt{lor(in(x,A), in(x,B))} is identical to \texttt{in(x,union(A,B))}).
Our full list of tasks can be found in~\cref{tab:atomic_tasks,,tab:all_tasks_word_comp,,tab:all_tasks_seq_comp,,tab:all_tasks_seq_comp_2}.

\paragraph{Sequential compositions}
Sequential composition functions convert word-wise tasks to sequence-level tasks. 
We specifically consider only two sequential functions: \texttt{map} and \texttt{filter}.
Note that %
compositions of multiple \texttt{map}s or multiple \texttt{filter}s can instead be expressed as compositions of multiple word-level functions. For example, 
\begin{center}
\texttt{map\{$\lambda$x.occupation(x)\}(map\{$\lambda$x. father(x)\}(S))}
\end{center}
(for an input sequence \texttt{S}) is equivalent to
\begin{center}
\texttt{map\{$\lambda$x.occupation(father(x))\}(S)}
\end{center}

Specifically, we define the following top-level sequential operator 
\begin{align}\label{eq:seq_formula}
\begin{split}
\texttt{map-}&\texttt{filter}\{f_M, f_F\} \\
& = \texttt{map}\{f_M\}(\texttt{filter}\{f_F\})
\end{split}
\end{align}
where $f_M$ is a word-wise relation and $f_F$ is a word-wise predicate. All recursively-defined sequential operators follow this form.
The following are the recursive rules for mapping nested maps and filters into a function of this form:
in the base cases,
\begin{align}
\begin{split}
\texttt{map}\{f_M\} &= \texttt{map-filter}\{f_M, \lambda\texttt{x.}\mathtt{true}\} \\
\texttt{filter}\{f_F\} &= \texttt{map-filter}\{\lambda\texttt{x.x}, f_F\}
;
\end{split}
\end{align}
in the recursive cases,
\begin{align}
\begin{split}
    \texttt{map}\{&f_M'\}(\texttt{map-filter}\{f_M, f_F\}) \\
    &= \texttt{map-filter}\{f_M'(f_M), f_F\} \\
    \texttt{filter}\{&f_F'\}(\texttt{map-filter}\{f_M, f_F\}) \\
    &= \texttt{map-filter}\{f_M, f_F\land f_F'(f_M)\}
.
\end{split}
\end{align}

\subsection{Dataset creation details}
\label{sec:app_methods_datasets}
Note that many tasks created through composition will be degenerate or identical to other tasks, even with our heuristic filters. We do a preliminary filter for degenerate tasks by removing tasks for which we have less than 100 samples. We also manually inspect all depth-2 word-level lexical compositions to ensure they are nontrivial and unique.

\paragraph{Word-level lexical tasks}
For English lexical tasks, we use words that appeared more than 5 times in the Brown corpus~\cite{brown} as our inputs $x$. For Spanish lexical tasks, we in use words that appeared at least once in the CESS Spanish Treebank~\cite{cess} as our inputs. This results in a a total of 9143 English words and 5298 Spanish words. %
We then construct outputs for each input word using either WordNet or SentiWordNet.
From each task, we filter out samples for which the relations map to an empty set---thus, for a task like \texttt{intersection(synonym(x), antonym(x))}, most samples will be filtered out as the set of synonyms are usually disjoint from the set of antonyms.
(This task ends up getting filtered out entirely, as the final number of samples is under 100.)

\paragraph{Word-level factual tasks}
We use a dump of Wikidata from 2017, taken from~\citet{sorokin-gurevych-2018-modeling}.\footnote{\url{https://public.ukp.informatik.tu-darmstadt.de/wikidata-dump/wikidata-virtuoso-dump-2017.zip}}
We convert each word-level factual task into SPARQL queries which returns a set of input-output data pairs from Wikidata.

For factual relations $R$, we create two queries: a \textit{sample} query which gives us a set of entities that participate in the relation, from which the inputs $x$ are derived, and a \textit{function} query that maps specific inputs $x$ to its set of output entities $R(x)$. For factual predicates $P$, we create three queries: a \textit{positive sample} query which gives samples $x$ for which $P(x) = \mathtt{true}$, a \textit{negative sample} query which gives samples $x$ for which $P(x) = \mathtt{false}$, and a \textit{function} query that maps specific inputs $x$ to its output boolean value $P(x)$.

The SPARQL query is generated recursively given the specification of the task. We define a function \texttt{task2sparql(T(x),y)} which converts tasks \texttt{T(x)} to SPARQL fragments (where the second argument to the function is the variable name we define for the output). We then convert the output of this function into a well-formed query using:
\begin{verbatim}
    SELECT ?x
    WHERE <task2sparql(T(x),y)>
\end{verbatim}
for sample queries and
\begin{verbatim}
    SELECT ?y
    WHERE <task2sparql(T(x),y)>
\end{verbatim}
for function queries. Note for function queries that the input \verb|x| is provided to us (and is not a variable).

The general rules specifying the \texttt{task2sparql} function are given in~\cref{tab:sparql_formula}.

\paragraph{Sequential tasks}
In practice, naively concatenating outputs from a random word sampler to create sequences will return degenerate or trivial sequences for many functions (for example, \texttt{map\{$\lambda$x. child(x)\}} is not meaningful for sequences consisting of words that don't refer to humans). Thus, we define a sequence sampler in~\cref{alg:seq_sampler} that takes in a sequential function (given in the form from~\cref{eq:seq_formula}), an input length $n$ and an output length $m\le n$, which will always sample sequences with length $n$ such that the output, when the function is applied to the sequence, is of length $m$.

\begin{algorithm}
\caption{Algorithm for sampling meaningful input sequences for sequential tasks.}
\label{alg:seq_sampler}
\small
\SetKwProg{Fn}{function}{:}{}
\Fn{seq\_sampler(\texttt{map-filter($f_M$,$f_F$)}, $n$, $m$)}{
    $seq\gets$``''\;
    \For{$i = 1\cdots n$}{
        $word\sim \mathtt{Unif}(\mathtt{domain}(f_M) \cap \{x: f_F(x) = \mathtt{true}\})$\;
        $seq\gets seq + word$
    }
    \For{$j = n\cdots m$}{
        $word\sim \mathtt{Unif}(\{x: f_F(x) = \mathtt{false}\})$\;
        $seq\gets seq + word$
    }
    $seq\gets\texttt{permute-words}(seq)$
}
\end{algorithm}

At a high level, this algorithm samples $n$ input words which are in the domain of the map relation, and for which the filter predicate returns $\mathtt{true}$, and $m-n$ input words for which the filter predicate returns $\mathtt{false}$, then permutes and concatenates them.

\begin{figure*}
    \centering
    \includegraphics[scale=0.4]{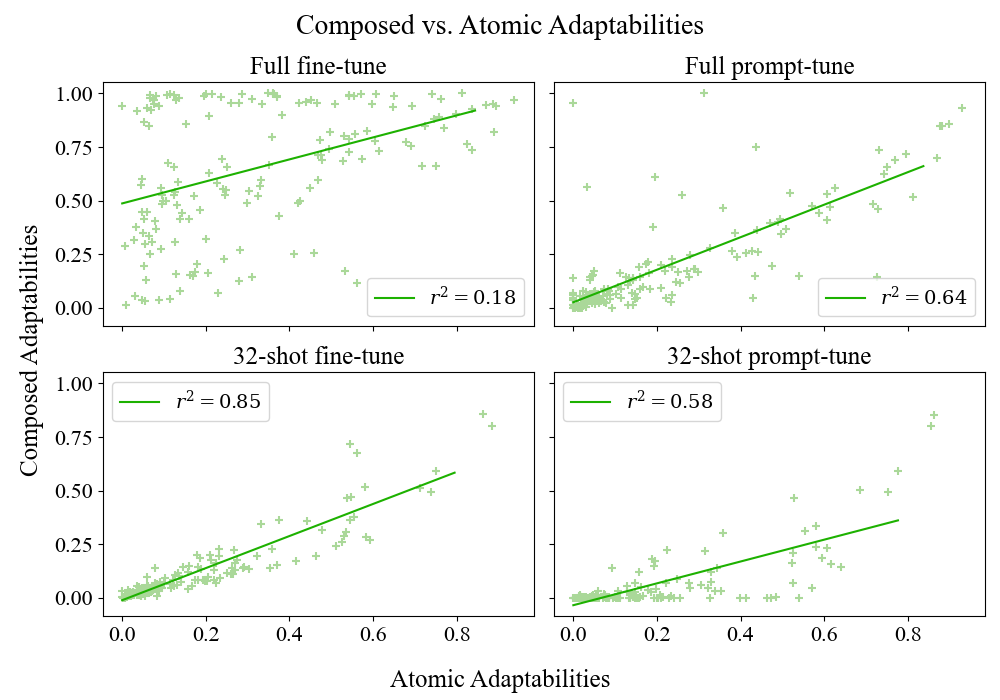}
    \caption{Compositionality of \texttt{map} function, when token separators are explicitly provided in the input and output. All adaptation paradigms demonstrate compositionality except for full fine-tuning, where there seems to be a large proportion of tasks for which the model can adapt to sequentially but not atomically.}
    \label{fig:composition_septokens}
\end{figure*}

\section{Experimental Setup Details}
\label{sec:appendix_training_details}
\paragraph{Hyperparameters}
We adapt a pre-trained T5-base model (24-layer, 220M parameters) to our tasks. 
We use an AdamW optimizer with a learning rate of 1.0 for all prompt-tuning experiments, and learning rate of 1e-3 for all fine-tuning experiments. We use batch sizes of 64 for word-level tasks, and 32 for sequential tasks. We run all experiments up to 100 epochs, and run 3--4 trials for each few-shot experiment to estimate average performance over possible choices of few-shot training samples.
These hyperparameters were chosen by trial and error on top of default configurations.

\paragraph{Infrastructure and Reproducibility}
For each task, we adapt our model using a single 32GB NVIDIA V100 GPU, or a single 40GB NVIDIA A100 GPU.
Training time varies by training dataset size and maximum number of epochs, but on average (using the hyperparameters specified above) is less than a few hours per task. %
Prompt-tuning is also more efficient than fine-tuning, updating the parameters of only 100 prompt tokens vs. the full 220M parameters in the model.

\paragraph{Evaluation of Sequential Tasks}
\label{sec:segmentation_seq}
When evaluating accuracies of sequential tasks (\cref{eq:accuracy}), note that we must align words in the generated sequence $y_i'$ with words in the ground-truth sequence $y_i$. However, this can be nontrivial, especially under the setting where word and entity boundaries are not explicitly generated by the model.
We cannot rely on whitespaces to segment words as a single word can span multiple white-spaces; for example, an entity \texttt{Will Smith} constitutes a single word. Instead, given a ground-truth sequence of $n$ words (note ground-truth segmentations are present in the dataset), we optimize over \text{all} possible length-$n$ segmentations of the generated sequence. %

\section{Compositionality Experiment: Additional Results}
\label{sec:app_comp}
Additional results for the compositionality experiment, including all composition functions, and the formula for the best-fit regression line in each case, %
are reported in~\cref{tab:composition}.
Furthermore, the \texttt{map} task with explicit segmentation (+separators) is plotted in isolation in~\cref{fig:composition_septokens}.

\section{Prediction distribution experiment: Additional details}
\label{sec:app_union}
We adapt the model to the task $\cup(f_e, f_h)$, %
constructing the training dataset for $\cup(f_e, f_h)$ to be balanced --- such that the model sees an equal number of examples of form $(x, f_e(x))$ as $(x, f_h(x))$.

Let $\mathcal{M}_{\cup(f_e,f_h)}$ denote a model adapted to this task.
Note that the domains of either function are not always identical, for example the set of entities in the domain of \texttt{political-party-of(x)} (mostly politicians) is different from the set of entities in the domain of \texttt{position-played-on-sports-team(x)} (mostly athletes). We create a balanced training set by first taking all items in the intersection of both domains, then sampling an equal number number of items in either domain.
Furthermore, to minimize the effect of the order seen during training, we shuffle the entire dataset after creating all example-label pairs. %

During test-time, we give $\mathcal{M}_{\cup(f_e,f_h)}$ a novel input $x'$ and 
record the average probability mass it assigned to all $y_e^i\in f_e(x')$ vs. all $y_h^i\in f_h(x')$. 
Note we evaluate only on inputs $x'$ which are in the domain of both $f_e$ and $f_h$.
Under the rare scenario that a prediction is in \textit{both} target tasks for a particular word (i.e. $y$ is in both $f_e(x')$ and $f_h(x')$), we count that towards both tasks, and increment the probability mass on both tasks by the probability the model assigned to $y$.

Instead of averaging across outputs in either set $f_e(x'),f_h(x')$, we also looked at the probabilities assigned to \textit{highest}-scoring predictions from each set. The overall trends were similar: the model tends to assign greater mass to the highest-scoring prediction from the easier task compared to highest-scoring prediction from the harder task.

\begin{figure*}
    \centering
    \includegraphics[scale=0.27,trim={0.5cm 3cm 9cm 0.4cm},clip]{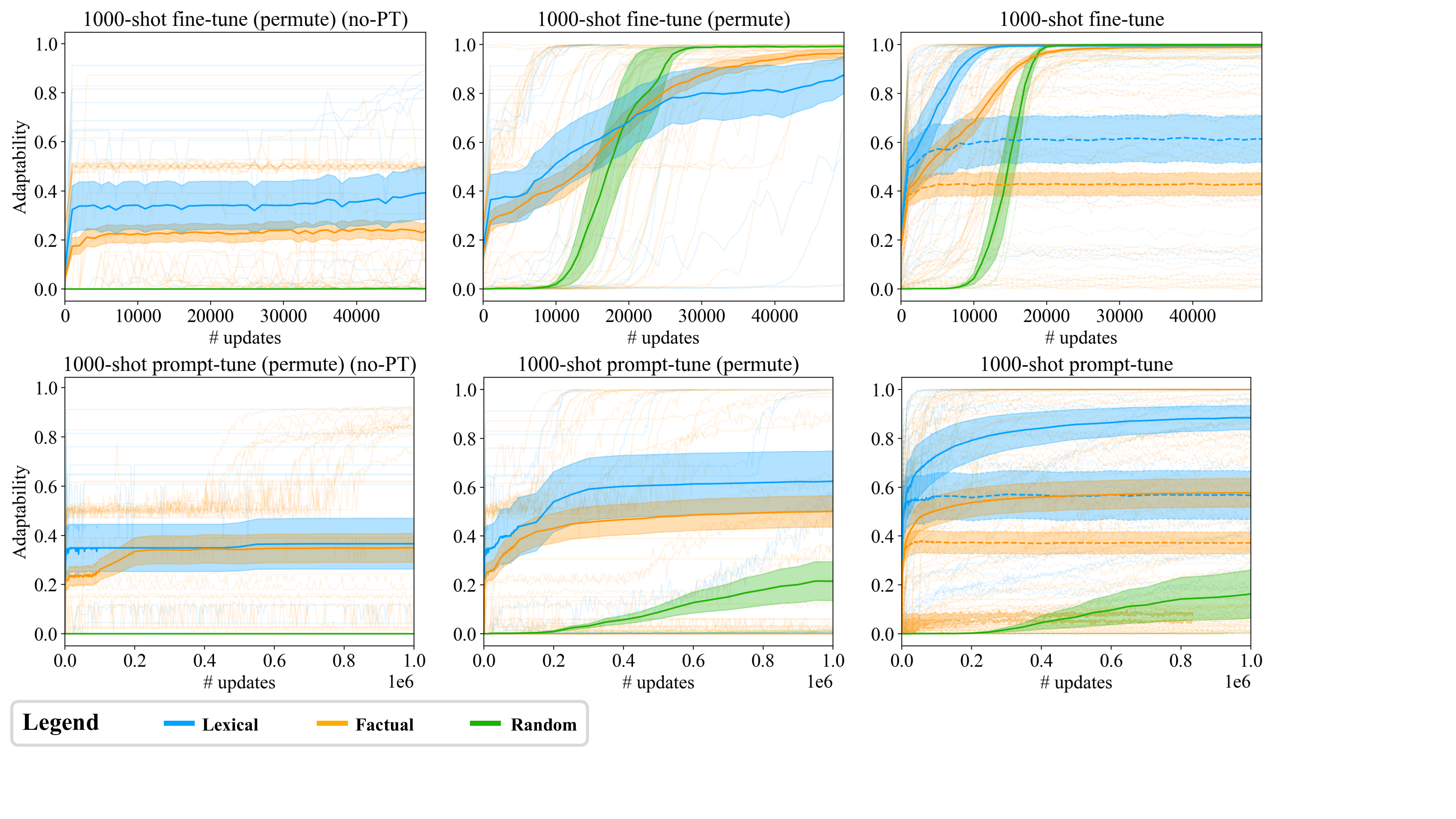}
    \caption{Memorization experiments on permuted vs. non-permuted versions of tasks, using pre-trained vs. non-pretrained models. Left figure shows an averaged memorization curve for a non-pretrained model on permuted tasks. Middle figure shows a pre-trained model on permuted tasks. Right figure shows a pre-trained model on non-permuted tasks. Pre-training enables models to adapt to novel tasks, but adapting to existing, non-permuted tasks is easier than adapting to novel, permuted tasks.}
    \label{fig:permute}
\end{figure*}
\section{Permuting task labels: disentangling effect of ``learning'' vs. ``retrieval''}
\label{sec:app_permute}
We hypothesize two ways that pre-trained models might adapt to new tasks: (1) through learning the underlying rules and patterns governing the task, or (2) through learning how to ``retrieve'' the correct label from memorized pre-training data.
These hypotheses, respectively, suggest two different roles for pre-training: (1) providing a ``generally good'' initialization from which many different tasks can be learned, or (2) imbuing the LM with memorized knowledge that can later be retrieved.

To determine which effect is at play (for which types of tasks),
for each atomic task, we permute the labels associated with each input, then run each adaptation paradigm on the permuted version of the task.
Notably, permuted labels differ from \texttt{random} tasks as the input and label distributions are restricted to be identical to original task.
Because the model would be unable to generalize to permuted labels, we only look at memorization ability.
The setting is similar to~\cref{sec:memorization}.
We compare the rate of adaptation for (A) a non-pretrained model to a permuted task, (B) a pre-trained model to a permuted task, and (C) a pre-trained model to non-permuted task.
If a pre-trained model is better able to adapt to a task than the non-pretrained model (B $>$ A)
, this indicates that pre-training helps models learn new tasks on the fly, supporting hypothesis 1.
If a pre-trained model can better adapt to a non-permuted task than it can to a permuted task (C $>$ B),
this indicates that adaptation requires something learned during pre-training, supporting hypothesis 2.

Results are shown in~\cref{fig:permute} (which, from left to right, shows settings A-C respectively). We find that for fine-tuning and prompt-tuning, both hypotheses are supported. For both lexical and factual tasks, pre-trained models can memorize novel word relations %
faster than non-pre-trained models.
However, pre-trained models can still adapt to non-permuted tasks faster than permuted ones.
Furthermore, note that for fine-tuning, the order of convergence of the three task types is reversed when going from permuted tasks to non-permuted tasks. In particular, random relations are easier to learn than permuted lexical or factual tasks. This suggests that \textit{models can more easily to adapt to random labels than labels that are known to be false}.

\begin{table*}[]
    \centering
    \begin{tabular}{ccccc}
    \toprule
        \textbf{Function type} & \textbf{Training type} & \textbf{Avg. adaptability} & \textbf{Optimal formula} & $r^2$ \textbf{value} \\
    \midrule
     & Full Fine-tuning & $37.43_{\pm 3.18}$ & $1.27x+0.14$ & $0.56$ \\
    Chaining & Full Prompt-tuning & $22.37_{\pm 3.03}$ & $1.32x+0.05$ & $0.65$ \\
    $f_2(f_1)$ & 32-shot Fine-tuning & $18.59_{\pm 2.21}$ & $1.34x+0.07$ & $0.57$ \\
     & 32-shot Prompt-tuning & $18.19_{\pm 2.21}$ & $1.32x+0.07$ & $0.6$ \\
    \midrule
     & Full Fine-tuning & $31.18_{\pm 2.02}$ & $1.24x+0.02$ & $0.73$ \\
    Union & Full Prompt-tuning & $25.05_{\pm 2.11}$ & $1.4x-0.01$ & $0.83$ \\
    $f_2\cup f_1$ & 32-shot Fine-tuning & $17.28_{\pm 1.52}$ & $1.37x+0.02$ & $0.8$ \\
     & 32-shot Prompt-tuning & $18.43_{\pm 1.55}$ & $1.35x+0.02$ & $0.8$ \\
    \midrule
     & Full Fine-tuning & $43.31_{\pm 22.42}$ & $2.25x-0.12$* & $0.97$* \\
    Intersection & Full Prompt-tuning & $16.68_{\pm 8.78}$ & $1.64x-0.04$* & $0.98$* \\
    $f_2\cap f_1$ & 32-shot Fine-tuning & $22.77_{\pm 17.03}$ & $5.93x-0.12$* & $0.91$* \\
     & 32-shot Prompt-tuning & $25.91_{\pm 19.38}$ & $6.81x-0.12$* & $0.94$* \\
    \midrule
     & Full Fine-tuning & $78.39_{\pm 2.53}$ & $2.15x-0.85$* & $0.8$* \\
    Logical And & Full Prompt-tuning & $79.25_{\pm 2.57}$ & $1.27x-0.18$* & $0.58$* \\
    $f_1\land f_2$ & 32-shot Fine-tuning & $66.49_{\pm 2.55}$ & $4.75x-2.13$* & $0.88$* \\
     & 32-shot Prompt-tuning & $55.86_{\pm 1.22}$ & $0.48x+0.3$* & $0.05$* \\
    \midrule
     & Full Fine-tuning & $72.41_{\pm 1.97}$ & $1.39x-0.37$* & $0.54$* \\
    Logical Or & Full Prompt-tuning & $74.71_{\pm 2.01}$ & $1.15x-0.18$* & $0.48$* \\
    $f_1\lor f_2$ & 32-shot Fine-tuning & $58.04_{\pm 1.11}$ & $1.52x-0.35$* & $0.63$* \\
     & 32-shot Prompt-tuning & $53.91_{\pm 0.48}$ & $0.8x+0.1$* & $0.33$* \\
    \midrule
     & Full Fine-tuning & $13.44_{\pm 1.73}$ & $0.15x+0.09$ & $0.03$ \\
    Map & Full Prompt-tuning & $5.39_{\pm 0.93}$ & $0.13x+0.03$ & $0.07$ \\
    \texttt{map\{$\lambda x. f_M(x)$\}} & 32-shot Fine-tuning & $3.59_{\pm 0.70}$ & $0.21x+0.0$ & $0.2$ \\
     & 32-shot Prompt-tuning & $3.77_{\pm 0.85}$ & $0.3x-0.01$ & $0.29$ \\
    \midrule
     & Full Fine-tuning & $67.40_{\pm 2.51}$ & $0.49x+0.52$ & $0.17$ \\
    Map (+separators) & Full Prompt-tuning & $18.02_{\pm 1.96}$ & $0.83x+0.02$ & $0.64$ \\
    \texttt{map\{$\lambda x. f_M(x)$\}} & 32-shot Fine-tuning & $10.66_{\pm 1.34}$ & $0.79x-0.01$ & $0.86$ \\
     & 32-shot Prompt-tuning & $5.22_{\pm 1.14}$ & $0.57x-0.04$ & $0.64$ \\
    \midrule
     & Full Fine-tuning & $82.08_{\pm 5.92}$ & $1.59x-0.58$* & $0.95$* \\
    Filter & Full Prompt-tuning & $78.58_{\pm 5.43}$ & $1.38x-0.43$* & $0.95$* \\
    \texttt{filter\{$\lambda x. f_F(x)$\}} & 32-shot Fine-tuning & $38.39_{\pm 3.27}$ & $0.81x-0.24$* & $0.87$* \\
     & 32-shot Prompt-tuning & $51.58_{\pm 4.99}$ & $1.19x-0.43$* & $0.87$* \\
    \bottomrule
    \end{tabular}
    \caption{We study the correlation between the atomic word-level functions and their compositions, under various training paradigms. We train a linear regressor to predict a model's generalization adaptability on a composite function based on its adaptabilities on the atomic constituents. Finally, we report the average generalization adaptability of composite tasks, for each training paradigm, under each type of composition. \\
    \small{* indicates composition function has less than 20 tasks, thus reported numbers may not be significant.}}
    \label{tab:composition}
\end{table*}
\begin{landscape}
\begin{table}[]
    \centering
    \tiny
    \begin{tabular}{clll}
    \toprule
    \textbf{Category} & \textbf{Predicates} & \textbf{Relations} \\
    \midrule
    Lexical
    & \verb|is-POS-noun[eng]| & \verb|synonyms[eng]| & \verb|synonyms[spa]| \\
    & \verb|is-POS-verb[eng]| & \verb|antonyms[eng]| & \verb|antonyms[spa]| \\
    & \verb|is-POS-adjective[eng]| & \verb|hyponyms[eng]|  & \verb|hyponyms[spa]| \\
    & \verb|is-POS-adverb[eng]| & \verb|entailments[eng]|  & \verb|entailments[spa]| \\
    & \verb|is-sentiment-positive[eng]| & \verb|translate[eng->spa]|  & \verb|translate[spa->eng]| \\
    & \verb|is-sentiment-negative[eng]| & \\
    & \verb|is-sentiment-neutral[eng]| & \\
    \midrule
    Factual
    & \verb|is-instance-human| & \verb|child| & \verb|location[inv]| \\
    & \verb|is-instance-film| & \verb|child[inv]| & \verb|manufacturer| \\
    & \verb|is-instance-book| & \verb|continent| & \verb|member of political party| \\
    & \verb|is-instance-city| & \verb|country of citizenship| & \verb|member of sports team| \\
    & \verb|is-instance-taxon| & \verb|country of origin| & \verb|mother| \\
    & \verb|is-occupation-actor| & \verb|country| & \verb|mother[inv]| \\
    & \verb|is-occupation-politician| & \verb|creator| & \verb|named after| \\
    & \verb|is-occupation-writer| & \verb|creator[inv]| & \verb|native language| \\
    & \verb|is-occupation-journalist| & \verb|developer| & \verb|occupation| \\
    & \verb|is-occupation-teacher| & \verb|diplomatic relation| & \verb|official language| \\
    & \verb|is-occupation-composer| & \verb|father| & \verb|original language of film or TV show| \\
    & \verb|is-birthplace-london| & \verb|father[inv]| & \verb|owned by| \\
    & \verb|is-birthplace-nyc| & \verb|genre| & \verb|performer| \\
    & \verb|is-birthplace-la| & \verb|has part| & \verb|place of birth| \\
    & \verb|is-birthplace-buenosaires| & \verb|head of state| & \verb|place of death| \\
    &  & \verb|head of state[inv]| & \verb|position held| \\
    &  & \verb|influenced by| & \verb|position played on team| \\
    &  & \verb|languages spoken written or signed| & \verb|record label| \\
    &  & \verb|location| & \verb|sex or gender| \\
    \midrule
    Random
    &  & \verb|random-seed0[eng]| & \verb|random-seed2[eng]| \\
    &  & \verb|random-seed1[eng]| & \verb|random-seed3[eng]| \\
    \bottomrule
    \end{tabular}
    \caption{Full list of atomic tasks in \ourdataset. %
    The content inside brackets specifies task input and output languages (\texttt{eng} for English and \texttt{spa} for Spanish). \texttt{\{inv\}} indicates the task is inverted, e.g. \texttt{creator} takes creations as input and returns their creators, while \texttt{creator\{inv\}} takes creators as input and returns their creations.}
    \label{tab:atomic_tasks}
\end{table}

\begin{table}[]
    \centering
    \tiny
    \begin{tabular}{rlll}
    \toprule
    \textbf{Category} & \textbf{Function} & \textbf{Example Tasks} & \textbf{Example Data} \\
    \midrule
    Chaining & \texttt{chain} & \texttt{mother(head of state)} & Russia $\to$ \{Maria Ivanovna Putina\} \\
    \midrule
    \multirow{2}{*}{Set Operations}
    & \texttt{union} & \texttt{union(mother, father)} & Elizabeth I of England $\to$ \{Anne Boleyn, Henry VIII of England\} \\
    & \texttt{intersection} & \texttt{intersection(entailments[eng], synonyms[eng])} & live $\to$ \{be, exist\} \\
    \midrule
    \multirow{7}{*}{Logical Operations}
    & \texttt{land} & \texttt{land(is-occupation-actor, is-birthplace-nyc)} & Anne Hathaway $\to \mathtt{true}$ \\
    &  &  & Brad Pitt $\to \mathtt{false}$ \\
    &  &  & Franklin Delano Roosevelt $\to \mathtt{false}$ \\
    \\
    & \texttt{lor} & \texttt{lor(is-birthplace-london, is-birthplace-nyc)} &  Franklin Delano Roosevelt $\to \mathtt{true}$ \\
    &  &  & David Bekham $\to\mathtt{true}$ \\
    &  &  & Mao Zedong $\to\mathtt{false}$ \\
    \midrule
    \multirow{2}{*}{Sequential Operations}
    & \texttt{map} & \texttt{map\{$\lambda$x. synonyms[eng]\}(S)} & criticality pillow delinquent culture eternity cling sane sentry $\to$ \{$\cdots$, criticalness rest neglectful acculturation timelessness cohere reasonable spotter, $\cdots$\} \\
    & \texttt{filter} & \texttt{$\lambda$x. filter\{is-POS-noun[eng]\}(S)} & expect inexpensive direct bones sullen breed switching eight $\to$ \{bones breed switching eight\} \\
    \bottomrule
    \end{tabular}
    \caption{Full list of composition functions used in \ourdataset, with examples.}
    \label{tab:comp_functions}
\end{table}
\end{landscape}
\begin{landscape}
\begin{table}{}
    \tiny
    \begin{tabularx}{\linewidth}{lllll}
    \toprule
    \verb|antonyms[eng](entailments[eng])| & \verb|entailments[spa](antonyms[spa])| & \verb|influenced by(creator)| & \verb|mother(creator)| & \verb|place of birth(influenced by)| \\
    \verb|antonyms[eng](hyponyms[eng])| & \verb|entailments[spa](hyponyms[spa])| & \verb|influenced by(father)| & \verb|mother(father)| & \verb|place of birth(named after)| \\
    \verb|antonyms[eng](translate[spa->eng])| & \verb|father(creator)| & \verb|influenced by(influenced by)| & \verb|mother(head of state)| & \verb|place of death(influenced by)| \\
    \verb|antonyms[spa](hyponyms[spa])| & \verb|father(father)| & \verb|influenced by(performer)| & \verb|mother(influenced by)| & \verb|place of death(named after)| \\
    \verb|child(influenced by)| & \verb|father(head of state)| & \verb|languages spoken written or signed(child)| & \verb|mother(mother)| & \verb|position held(influenced by)| \\
    \verb|child(named after)| & \verb|father(mother)| & \verb|languages spoken written or signed(influenced by)| & \verb|mother(named after)| & \verb|position held(mother)| \\
    \verb|child(owned by)| & \verb|father(named after)| & \verb|languages spoken written or signed(named after)| & \verb|mother(owned by)| & \verb|position played on team(father)| \\
    \verb|country of citizenship(child)| & \verb|hyponyms[eng](antonyms[eng])| & \verb|member of political party(father)| & \verb|mother(performer)| & \verb|position played on team(named after)| \\
    \verb|country of citizenship(father)| & \verb|hyponyms[eng](entailments[eng])| & \verb|member of political party(influenced by)| & \verb|named after(child)| & \verb|record label(child)| \\
    \verb|country of citizenship(mother)| & \verb|hyponyms[eng](translate[spa->eng])| & \verb|member of political party(mother)| & \verb|named after(developer)| & \verb|record label(father)| \\
    \verb|entailments[eng](antonyms[eng])| & \verb|hyponyms[spa](antonyms[spa])| & \verb|member of political party(named after)| & \verb|named after(influenced by)| & \verb|record label(influenced by)| \\
    \verb|entailments[eng](hyponyms[eng])| & \verb|hyponyms[spa](entailments[spa])| & \verb|member of sports team(child)| & \verb|occupation(influenced by)| & \verb|record label(mother)| \\
    \verb|entailments[eng](translate[spa->eng])| & \verb|influenced by(child)| & \verb|member of sports team(father)| & \verb|occupation(named after)| & \verb|translate[eng->spa](antonyms[eng])| \\
    \verb|entailments[spa](antonyms[spa])| & \verb|influenced by(creator)| \\
    \\
    \midrule
    \end{tabularx}
    
    \begin{tabularx}{\linewidth}{XXX}
    \\
    \verb|union(antonyms[eng], entailments[eng])| & \verb|union(father, mother)| & \verb|union(place of death, position held)| \\
    \verb|union(antonyms[eng], hyponyms[eng])| & \verb|union(hyponyms[eng], synonyms[eng])| & \verb|union(place of death, position played on team)| \\
    \verb|union(antonyms[eng], synonyms[eng])| & \verb|union(hyponyms[spa], synonyms[spa])| & \verb|union(place of death, record label)| \\
    \verb|union(antonyms[spa], entailments[spa])| & \verb|union(languages spoken written or signed, member of political party)| & \verb|union(random-seed0[eng], antonyms[eng])| \\
    \verb|union(antonyms[spa], hyponyms[spa])| & \verb|union(languages spoken written or signed, mother)| & \verb|union(random-seed0[eng], entailments[eng])| \\
    \verb|union(antonyms[spa], synonyms[spa])| & \verb|union(languages spoken written or signed, occupation)| & \verb|union(random-seed0[eng], hyponyms[eng])| \\
    \verb|union(child, father)| & \verb|union(languages spoken written or signed, place of birth)| & \verb|union(random-seed0[eng], synonyms[eng])| \\
    \verb|union(child, mother)| & \verb|union(languages spoken written or signed, position held)| & \verb|union(random-seed1[eng], antonyms[eng])| \\
    \verb|union(country of citizenship, languages spoken written or signed)| & \verb|union(languages spoken written or signed, position played on team)| & \verb|union(random-seed1[eng], entailments[eng])| \\
    \verb|union(country of citizenship, named after)| & \verb|union(languages spoken written or signed, record label)| & \verb|union(random-seed1[eng], hyponyms[eng])| \\
    \verb|union(country of citizenship, position held)| & \verb|union(member of political party, mother)| & \verb|union(random-seed1[eng], synonyms[eng])| \\
    \verb|union(country of citizenship, position played on team)| & \verb|union(member of political party, place of birth)| & \verb|union(random-seed2[eng], antonyms[eng])| \\
    \verb|union(creator, father)| & \verb|union(member of political party, record label)| & \verb|union(random-seed2[eng], entailments[eng])| \\
    \verb|union(creator, mother)| & \verb|union(member of sports team, mother)| & \verb|union(random-seed2[eng], hyponyms[eng])| \\
    \verb|union(entailments[eng], hyponyms[eng])| & \verb|union(member of sports team, place of death)| & \verb|union(random-seed2[eng], synonyms[eng])| \\
    \verb|union(entailments[eng], synonyms[eng])| & \verb|union(occupation, place of death)| & \verb|union(random-seed3[eng], antonyms[eng])| \\
    \verb|union(entailments[spa], hyponyms[spa])| & \verb|union(place of birth, position held)| & \verb|union(random-seed3[eng], entailments[eng])| \\
    \verb|union(entailments[spa], synonyms[spa])| & \verb|union(place of birth, position played on team)| & \verb|union(random-seed3[eng], hyponyms[eng])| \\
    \verb|union(father, influenced by)| & \verb|union(place of birth, record label)| & \verb|union(random-seed3[eng], synonyms[eng])| \\
    \\
    \midrule
    \end{tabularx}
    
    \begin{tabularx}{\linewidth}{XXX}
    \\
    \verb|intersection(entailments[eng], synonyms[eng])| & \verb|intersection(hyponyms[eng], synonyms[eng])| & \verb|intersection(hyponyms[spa], synonyms[spa])| \\
    \\
    \midrule
    \end{tabularx}
    
    \begin{tabularx}{\linewidth}{XXX}
    \\
    \verb|land(is-occupation-actor, is-birthplace-buenosaires)| & \verb|land(is-occupation-actor, is-birthplace-london)| & \verb|land(is-occupation-politician, is-birthplace-london)| \\
    \verb|land(is-occupation-actor, is-birthplace-la)| & \verb|land(is-occupation-actor, is-birthplace-nyc)| & \verb|land(is-occupation-politician, is-birthplace-nyc)| \\
    \midrule
    \end{tabularx}
    
    \begin{tabularx}{\linewidth}{llll}
    \\
    \verb|lor(is-birthplace-buenosaires, is-occupation-journalist)| & \verb|lor(is-birthplace-la, is-birthplace-london)| & \verb|lor(is-birthplace-nyc, is-birthplace-london)| & \verb|lor(is-occupation-actor, is-birthplace-la)| \\
    \verb|lor(is-birthplace-buenosaires, is-occupation-politician)| & \verb|lor(is-birthplace-london, is-occupation-teacher)| & \verb|lor(is-birthplace-nyc, is-occupation-actor)| & \verb|lor(is-occupation-actor, is-birthplace-nyc)| \\
    \verb|lor(is-birthplace-buenosaires, is-occupation-teacher)| & \verb|lor(is-birthplace-nyc, is-birthplace-buenosaires)| & \verb|lor(is-birthplace-nyc, is-occupation-politician)| & \verb|lor(is-occupation-journalist, is-birthplace-buenosaires)| \\
    \verb|lor(is-birthplace-la, is-birthplace-buenosaires)| & \verb|lor(is-birthplace-nyc, is-birthplace-la)| & \verb|lor(is-occupation-actor, is-birthplace-buenosaires)| & \verb|lor(is-occupation-politician, is-birthplace-buenosaires)| \\
    \verb|lor(is-birthplace-la, is-birthplace-london)| \\
    \bottomrule
    \end{tabularx}
    \caption{Full list of word-level compositional tasks in \ourdataset, organized by composition type.}
    \label{tab:all_tasks_word_comp}
\end{table}
\end{landscape}

\begin{landscape}
\begin{table}[]
    \tiny
    \begin{tabularx}{\linewidth}{lll}
    \toprule
    \texttt{map\{$\lambda$x. antonyms[eng](entailments[eng](x))\}} & \texttt{map\{$\lambda$x. languages spoken written or signed(child(x))\}} & \texttt{map\{$\lambda$x. translate[eng->spa](x)\}} \\
    \texttt{map\{$\lambda$x. antonyms[eng](hyponyms[eng](x))\}} & \texttt{map\{$\lambda$x. languages spoken written or signed(influenced by(x))\}} & \texttt{map\{$\lambda$x. translate[spa->eng](x)\}} \\
    \texttt{map\{$\lambda$x. antonyms[eng](translate[spa->eng](x))\}} & \texttt{map\{$\lambda$x. languages spoken written or signed(named after(x))\}} & \texttt{map\{$\lambda$x. union(antonyms[eng](x), entailments[eng](x))\}} \\
    \texttt{map\{$\lambda$x. antonyms[eng](x)\}} & \texttt{map\{$\lambda$x. languages spoken written or signed(x)\}} & \texttt{map\{$\lambda$x. union(antonyms[eng](x), hyponyms[eng](x))\}} \\
    \texttt{map\{$\lambda$x. antonyms[spa](entailments[spa](x))\}} & \texttt{map\{$\lambda$x. location(x)\}} & \texttt{map\{$\lambda$x. union(antonyms[eng](x), synonyms[eng](x))\}} \\
    \texttt{map\{$\lambda$x. antonyms[spa](hyponyms[spa](x))\}} & \texttt{map\{$\lambda$x. location[inv](x)\}} & \texttt{map\{$\lambda$x. union(antonyms[spa](x), entailments[spa](x))\}} \\
    \texttt{map\{$\lambda$x. antonyms[spa](x)\}} & \texttt{map\{$\lambda$x. manufacturer(x)\}} & \texttt{map\{$\lambda$x. union(antonyms[spa](x), hyponyms[spa](x))\}} \\
    \texttt{map\{$\lambda$x. child(influenced by(x))\}} & \texttt{map\{$\lambda$x. member of political party(father(x))\}} & \texttt{map\{$\lambda$x. union(antonyms[spa](x), synonyms[spa](x))\}} \\
    \texttt{map\{$\lambda$x. child(named after(x))\}} & \texttt{map\{$\lambda$x. member of political party(influenced by(x))\}} & \texttt{map\{$\lambda$x. union(child(x), father(x))\}} \\
    \texttt{map\{$\lambda$x. child(owned by(x))\}} & \texttt{map\{$\lambda$x. member of political party(mother(x))\}} & \texttt{map\{$\lambda$x. union(child(x), mother(x))\}} \\
    \texttt{map\{$\lambda$x. child[inv](x)\}} & \texttt{map\{$\lambda$x. member of political party(named after(x))\}} & \texttt{map\{$\lambda$x. union(child(x), named after(x))\}} \\
    \texttt{map\{$\lambda$x. continent(x)\}} & \texttt{map\{$\lambda$x. member of political party(x)\}} & \texttt{map\{$\lambda$x. union(country of citizenship(x), languages spoken written or signed(x))\}} \\
    \texttt{map\{$\lambda$x. country of citizenship(child(x))\}} & \texttt{map\{$\lambda$x. member of sports team(child(x))\}} & \texttt{map\{$\lambda$x. union(country of citizenship(x), named after(x))\}} \\
    \texttt{map\{$\lambda$x. country of citizenship(father(x))\}} & \texttt{map\{$\lambda$x. member of sports team(father(x))\}} & \texttt{map\{$\lambda$x. union(country of citizenship(x), position held(x))\}} \\
    \texttt{map\{$\lambda$x. country of citizenship(mother(x))\}} & \texttt{map\{$\lambda$x. member of sports team(influenced by(x))\}} & \texttt{map\{$\lambda$x. union(country of citizenship(x), position played on team(x))\}} \\
    \texttt{map\{$\lambda$x. country of citizenship(x)\}} & \texttt{map\{$\lambda$x. member of sports team(x)\}} & \texttt{map\{$\lambda$x. union(creator(x), father(x))\}} \\
    \texttt{map\{$\lambda$x. country of origin(x)\}} & \texttt{map\{$\lambda$x. mother(creator(x))\}} & \texttt{map\{$\lambda$x. union(creator(x), location(x))\}} \\
    \texttt{map\{$\lambda$x. country(x)\}} & \texttt{map\{$\lambda$x. mother(father(x))\}} & \texttt{map\{$\lambda$x. union(creator(x), mother(x))\}} \\
    \texttt{map\{$\lambda$x. creator(x)\}} & \texttt{map\{$\lambda$x. mother(head of state(x))\}} & \texttt{map\{$\lambda$x. union(entailments[eng](x), hyponyms[eng](x))}(S)map{child(x)\}} \\
    \texttt{map\{$\lambda$x. creator[inv](x)\}} & \texttt{map\{$\lambda$x. mother(influenced by(x))\}} & \texttt{map\{$\lambda$x. union(entailments[eng](x), synonyms[eng](x))\}} \\
    \texttt{map\{$\lambda$x. developer(x)\}} & \texttt{map\{$\lambda$x. mother(mother(x))\}} & \texttt{map\{$\lambda$x. union(entailments[spa](x), hyponyms[spa](x))\}} \\
    \texttt{map\{$\lambda$x. diplomatic relation(x)\}} & \texttt{map\{$\lambda$x. mother(named after(x))\}} & \texttt{map\{$\lambda$x. union(entailments[spa](x), synonyms[spa](x))\}} \\
    \texttt{map\{$\lambda$x. entailments[eng](antonyms[eng](x))\}} & \texttt{map\{$\lambda$x. mother(owned by(x))\}} & \texttt{map\{$\lambda$x. union(father(x), influenced by(x))\}} \\
    \texttt{map\{$\lambda$x. entailments[eng](hyponyms[eng](x))\}} & \texttt{map\{$\lambda$x. mother(performer(x))\}} & \texttt{map\{$\lambda$x. union(father(x), mother(x))\}} \\
    \texttt{map\{$\lambda$x. entailments[eng](translate[spa->eng](x))\}} & \texttt{map\{$\lambda$x. mother(x)\}} & \texttt{map\{$\lambda$x. union(father(x), named after(x))\}} \\
    \texttt{map\{$\lambda$x. entailments[eng](x)\}} & \texttt{map\{$\lambda$x. mother[inv](x)\}} & \texttt{map\{$\lambda$x. union(hyponyms[eng](x), synonyms[eng](x))\}} \\
    \texttt{map\{$\lambda$x. entailments[spa](antonyms[spa](x))\}} & \texttt{map\{$\lambda$x. named after(child(x))\}} & \texttt{map\{$\lambda$x. union(hyponyms[spa](x), synonyms[spa](x))\}} \\
    \texttt{map\{$\lambda$x. entailments[spa](hyponyms[spa](x))\}} & \texttt{map\{$\lambda$x. named after(creator(x))\}} & \texttt{map\{$\lambda$x. union(influenced by(x), mother(x))\}} \\
    \texttt{map\{$\lambda$x. entailments[spa](x)\}} & \texttt{map\{$\lambda$x. named after(developer(x))\}} & \texttt{map\{$\lambda$x. union(influenced by(x), named after(x))\}} \\
    \texttt{map\{$\lambda$x. father(creator(x))\}} & \texttt{map\{$\lambda$x. named after(father(x))\}} & \texttt{map\{$\lambda$x. union(languages spoken written or signed(x), member of political party(x))\}} \\
    \texttt{map\{$\lambda$x. father(father(x))\}} & \texttt{map\{$\lambda$x. named after(influenced by(x))\}} & \texttt{map\{$\lambda$x. union(languages spoken written or signed(x), member of sports team(x))\}} \\
    \texttt{map\{$\lambda$x. father(head of state(x))\}} & \texttt{map\{$\lambda$x. named after(x)\}} & \texttt{map\{$\lambda$x. union(languages spoken written or signed(x), mother(x))\}} \\
    \texttt{map\{$\lambda$x. father(mother(x))\}} & \texttt{map\{$\lambda$x. native language(x)\}} & \texttt{map\{$\lambda$x. union(languages spoken written or signed(x), occupation(x))\}} \\
    \texttt{map\{$\lambda$x. father(named after(x))\}} & \texttt{map\{$\lambda$x. occupation(influenced by(x))\}} & \texttt{map\{$\lambda$x. union(languages spoken written or signed(x), place of birth(x))\}} \\
    \texttt{map\{$\lambda$x. father(x)\}} & \texttt{map\{$\lambda$x. occupation(named after(x))\}} & \texttt{map\{$\lambda$x. union(languages spoken written or signed(x), position held(x))\}} \\
    \texttt{map\{$\lambda$x. father[inv](x)\}} & \texttt{map\{$\lambda$x. occupation(x)\}} & \texttt{map\{$\lambda$x. union(languages spoken written or signed(x), position played on team(x))\}} \\
    \texttt{map\{$\lambda$x. genre(x)\}} & \texttt{map\{$\lambda$x. official language(x)\}} & \texttt{map\{$\lambda$x. union(languages spoken written or signed(x), record label(x))\}} \\
    \texttt{map\{$\lambda$x. has part(x)\}} & \texttt{map\{$\lambda$x. original language of film or TV show(x)\}} & \texttt{map\{$\lambda$x. union(member of political party(x), mother(x))\}} \\
    \texttt{map\{$\lambda$x. head of state(x)\}} & \texttt{map\{$\lambda$x. owned by(x)\}} & \texttt{map\{$\lambda$x. union(member of political party(x), named after(x))\}} \\
    \texttt{map\{$\lambda$x. head of state[inv](x)\}} & \texttt{map\{$\lambda$x. performer(x)\}} & \texttt{map\{$\lambda$x. union(member of political party(x), place of birth(x))\}} \\
    \texttt{map\{$\lambda$x. hyponyms[eng](antonyms[eng](x))\}} & \texttt{map\{$\lambda$x. place of birth(influenced by(x))\}} & \texttt{map\{$\lambda$x. union(member of political party(x), record label(x))\}} \\
    \texttt{map\{$\lambda$x. hyponyms[eng](entailments[eng](x))\}} & \texttt{map\{$\lambda$x. place of birth(named after(x))\}} & \texttt{map\{$\lambda$x. union(member of sports team(x), mother(x))\}} \\
    \texttt{map\{$\lambda$x. hyponyms[eng](translate[spa->eng](x))\}} & \texttt{map\{$\lambda$x. place of birth(x)\}} & \texttt{map\{$\lambda$x. union(member of sports team(x), place of death(x))\}} \\
    \texttt{map\{$\lambda$x. hyponyms[eng](x)\}} & \texttt{map\{$\lambda$x. place of death(influenced by(x))\}} & \texttt{map\{$\lambda$x. union(member of sports team(x), record label(x))\}} \\
    \texttt{map\{$\lambda$x. hyponyms[spa](antonyms[spa](x))\}} & \texttt{map\{$\lambda$x. place of death(named after(x))\}} & \texttt{map\{$\lambda$x. union(mother(x), named after(x))\}} \\
    \texttt{map\{$\lambda$x. hyponyms[spa](entailments[spa](x))\}} & \texttt{map\{$\lambda$x. place of death(x)\}} & \texttt{map\{$\lambda$x. union(occupation(x), place of death(x))\}} \\
    \texttt{map\{$\lambda$x. hyponyms[spa](x)\}} & \texttt{map\{$\lambda$x. position held(influenced by(x))\}} & \texttt{map\{$\lambda$x. union(place of birth(x), position held(x))\}} \\
    \texttt{map\{$\lambda$x. influenced by(child(x))\}} & \texttt{map\{$\lambda$x. position held(mother(x))\}} & \texttt{map\{$\lambda$x. union(place of birth(x), position played on team(x))\}} \\
    \texttt{map\{$\lambda$x. influenced by(creator(x))\}} & \texttt{map\{$\lambda$x. position held(x)\}} & \texttt{map\{$\lambda$x. union(place of birth(x), record label(x))\}} \\
    \texttt{map\{$\lambda$x. influenced by(developer(x))\}} & \texttt{map\{$\lambda$x. position played on team(father(x))\}} & \texttt{map\{$\lambda$x. union(place of death(x), position held(x))\}} \\
    \texttt{map\{$\lambda$x. influenced by(father(x))\}} & \texttt{map\{$\lambda$x. position played on team(named after(x))\}} & \texttt{map\{$\lambda$x. union(place of death(x), position played on team(x))\}} \\
    \texttt{map\{$\lambda$x. influenced by(head of state(x))\}} & \texttt{map\{$\lambda$x. position played on team(x)\}} & \texttt{map\{$\lambda$x. union(place of death(x), record label(x))\}} \\
    \texttt{map\{$\lambda$x. influenced by(influenced by(x))\}} & \texttt{map\{$\lambda$x. random-seed0[eng](x)\}} & \texttt{map\{$\lambda$x. union(random-seed0[eng](x), antonyms[eng](x))\}} \\
    \texttt{map\{$\lambda$x. influenced by(owned by(x))\}} & \texttt{map\{$\lambda$x. random-seed1[eng](x)\}} & \texttt{map\{$\lambda$x. union(random-seed0[eng](x), entailments[eng](x))\}} \\
    \texttt{map\{$\lambda$x. influenced by(performer(x))\}} & \texttt{map\{$\lambda$x. random-seed2[eng](x)\}} & \texttt{map\{$\lambda$x. union(random-seed0[eng](x), hyponyms[eng](x))\}} \\
    \texttt{map\{$\lambda$x. influenced by(x)\}} & \texttt{map\{$\lambda$x. random-seed3[eng](x)\}} & \texttt{map\{$\lambda$x. union(random-seed0[eng](x), synonyms[eng](x))\}} \\
    \texttt{map\{$\lambda$x. intersection(antonyms[eng](x), entailments[eng](x))\}} & \texttt{map\{$\lambda$x. record label(child(x))\}} & \texttt{map\{$\lambda$x. union(random-seed1[eng](x), antonyms[eng](x))\}} \\
    \texttt{map\{$\lambda$x. intersection(antonyms[eng](x), hyponyms[eng](x))\}} & \texttt{map\{$\lambda$x. record label(father(x))\}} & \texttt{map\{$\lambda$x. union(random-seed1[eng](x), entailments[eng](x))\}} \\
    \texttt{map\{$\lambda$x. intersection(antonyms[eng](x), synonyms[eng](x))\}} & \texttt{map\{$\lambda$x. record label(influenced by(x))\}} & \texttt{map\{$\lambda$x. union(random-seed1[eng](x), hyponyms[eng](x))\}} \\
    \texttt{map\{$\lambda$x. intersection(antonyms[spa](x), entailments[spa](x))\}} & \texttt{map\{$\lambda$x. record label(mother(x))\}} & \texttt{map\{$\lambda$x. union(random-seed1[eng](x), synonyms[eng](x))\}} \\
    \texttt{map\{$\lambda$x. intersection(antonyms[spa](x), hyponyms[spa](x))\}} & \texttt{map\{$\lambda$x. record label(x)\}} & \texttt{map\{$\lambda$x. union(random-seed2[eng](x), antonyms[eng](x))\}} \\
    \texttt{map\{$\lambda$x. intersection(antonyms[spa](x), synonyms[spa](x))\}} & \texttt{map\{$\lambda$x. sex or gender(x)\}} & \texttt{map\{$\lambda$x. union(random-seed2[eng](x), entailments[eng](x))\}} \\
    \texttt{map\{$\lambda$x. intersection(entailments[eng](x), hyponyms[eng](x))\}} & \texttt{map\{$\lambda$x. subclass of(x)\}} & \texttt{map\{$\lambda$x. union(random-seed2[eng](x), hyponyms[eng](x))\}} \\
    \texttt{map\{$\lambda$x. intersection(entailments[eng](x), synonyms[eng](x))\}} & \texttt{map\{$\lambda$x. synonyms[eng](x)\}} & \texttt{map\{$\lambda$x. union(random-seed2[eng](x), synonyms[eng](x))\}} \\
    \texttt{map\{$\lambda$x. intersection(entailments[spa](x), hyponyms[spa](x))\}} & \texttt{map\{$\lambda$x. synonyms[spa](x)\}} & \texttt{map\{$\lambda$x. union(random-seed3[eng](x), antonyms[eng](x))\}} \\
    \texttt{map\{$\lambda$x. intersection(entailments[spa](x), synonyms[spa](x))\}} & \texttt{map\{$\lambda$x. translate[eng->spa](antonyms[eng](x))\}} & \texttt{map\{$\lambda$x. union(random-seed3[eng](x), entailments[eng](x))\}} \\
    \texttt{map\{$\lambda$x. intersection(hyponyms[eng](x), synonyms[eng](x))\}} & \texttt{map\{$\lambda$x. translate[eng->spa](entailments[eng](x))\}} & \texttt{map\{$\lambda$x. union(random-seed3[eng](x), hyponyms[eng](x))\}} \\
    \texttt{map\{$\lambda$x. intersection(hyponyms[spa](x), synonyms[spa](x))\}} & \texttt{map\{$\lambda$x. translate[eng->spa](hyponyms[eng](x))\}} & \texttt{map\{$\lambda$x. union(random-seed3[eng](x), synonyms[eng](x))\}} \\
    \bottomrule
    \end{tabularx}
    \caption{Full list of sequential compositional tasks in \ourdataset, organized by composition type.}
    \label{tab:all_tasks_seq_comp}
\end{table}
\end{landscape}

\begin{landscape}
\begin{table}[]
    \tiny
    \centering
    \begin{tabularx}{\linewidth}{XXXXXXX}
    \toprule
    \texttt{filter\{$\lambda$x. is-POS-adjective[eng](x)\}} & \texttt{filter\{$\lambda$x. is-POS-adverb[eng](x)\}} & \texttt{filter\{$\lambda$x. is-POS-noun[eng](x)\}} & \texttt{filter\{$\lambda$x. is-POS-verb[eng](x)\}} & \texttt{filter\{$\lambda$x. is-sentiment-negative[eng](x)\}} \\
    \texttt{filter\{$\lambda$x. is-POS-adverb[eng](x)\}} & \texttt{filter\{$\lambda$x. is-POS-noun[eng](x)\}} \\
    \\
    \midrule
    \end{tabularx}
    
    \begin{tabularx}{\linewidth}{llll}
    \\
    \texttt{map\{$\lambda$x. antonyms[eng](x)\}(filter\{$\lambda$x. is-POS-adjective[eng](x)\})} & \texttt{map\{$\lambda$x. random-seed0[eng](x)\}(filter\{$\lambda$x. is-POS-adjective[eng](x)\})} & \texttt{map\{$\lambda$x. random-seed3[eng](x)\}(filter\{$\lambda$x. is-POS-adjective[eng](x)\})} \\
    \texttt{map\{$\lambda$x. antonyms[eng](x)\}(filter\{$\lambda$x. is-POS-adverb[eng](x)\})} & \texttt{map\{$\lambda$x. random-seed0[eng](x)\}(filter\{$\lambda$x. is-POS-adverb[eng](x)\})} & \texttt{map\{$\lambda$x. random-seed3[eng](x)\}(filter\{$\lambda$x. is-POS-adverb[eng](x)\})} \\
    \texttt{map\{$\lambda$x. antonyms[eng](x)\}(filter\{$\lambda$x. is-POS-noun[eng](x)\})} & \texttt{map\{$\lambda$x. random-seed0[eng](x)\}(filter\{$\lambda$x. is-POS-noun[eng](x)\})} & \texttt{map\{$\lambda$x. random-seed3[eng](x)\}(filter\{$\lambda$x. is-POS-noun[eng](x)\})} \\
    \texttt{map\{$\lambda$x. antonyms[eng](x)\}(filter\{$\lambda$x. is-POS-verb[eng](x)\})} & \texttt{map\{$\lambda$x. random-seed0[eng](x)\}(filter\{$\lambda$x. is-POS-verb[eng](x)\})} & \texttt{map\{$\lambda$x. random-seed3[eng](x)\}(filter\{$\lambda$x. is-POS-verb[eng](x)\})} \\
    \texttt{map\{$\lambda$x. antonyms[eng](x)\}(filter\{$\lambda$x. is-sentiment-negative[eng](x)\})} & \texttt{map\{$\lambda$x. random-seed0[eng](x)\}(filter\{$\lambda$x. is-sentiment-negative[eng](x)\})} & \texttt{map\{$\lambda$x. random-seed3[eng](x)\}(filter\{$\lambda$x. is-sentiment-negative[eng](x)\})} \\
    \texttt{map\{$\lambda$x. antonyms[eng](x)\}(filter\{$\lambda$x. is-sentiment-neutral[eng](x)\})} & \texttt{map\{$\lambda$x. random-seed0[eng](x)\}(filter\{$\lambda$x. is-sentiment-neutral[eng](x)\})} & \texttt{map\{$\lambda$x. random-seed3[eng](x)\}(filter\{$\lambda$x. is-sentiment-neutral[eng](x)\})} \\
    \texttt{map\{$\lambda$x. antonyms[eng](x)\}(filter\{$\lambda$x. is-sentiment-positive[eng](x)\})} & \texttt{map\{$\lambda$x. random-seed0[eng](x)\}(filter\{$\lambda$x. is-sentiment-positive[eng](x)\})} & \texttt{map\{$\lambda$x. random-seed3[eng](x)\}(filter\{$\lambda$x. is-sentiment-positive[eng](x)\})} \\
    \texttt{map\{$\lambda$x. entailments[eng](x)\}(filter\{$\lambda$x. is-POS-adjective[eng](x)\})} & \texttt{map\{$\lambda$x. random-seed1[eng](x)\}(filter\{$\lambda$x. is-POS-adjective[eng](x)\})} & \texttt{map\{$\lambda$x. synonyms[eng](x)\}(filter\{$\lambda$x. is-POS-adjective[eng](x)\})} \\
    \texttt{map\{$\lambda$x. entailments[eng](x)\}(filter\{$\lambda$x. is-POS-adverb[eng](x)\})} & \texttt{map\{$\lambda$x. random-seed1[eng](x)\}(filter\{$\lambda$x. is-POS-adverb[eng](x)\})} & \texttt{map\{$\lambda$x. synonyms[eng](x)\}(filter\{$\lambda$x. is-POS-adverb[eng](x)\})} \\
    \texttt{map\{$\lambda$x. entailments[eng](x)\}(filter\{$\lambda$x. is-POS-noun[eng](x)\})} & \texttt{map\{$\lambda$x. random-seed1[eng](x)\}(filter\{$\lambda$x. is-POS-noun[eng](x)\})} & \texttt{map\{$\lambda$x. synonyms[eng](x)\}(filter\{$\lambda$x. is-POS-noun[eng](x)\})} \\
    \texttt{map\{$\lambda$x. entailments[eng](x)\}(filter\{$\lambda$x. is-POS-verb[eng](x)\})} & \texttt{map\{$\lambda$x. random-seed1[eng](x)\}(filter\{$\lambda$x. is-POS-verb[eng](x)\})} & \texttt{map\{$\lambda$x. synonyms[eng](x)\}(filter\{$\lambda$x. is-POS-verb[eng](x)\})} \\
    \texttt{map\{$\lambda$x. entailments[eng](x)\}(filter\{$\lambda$x. is-sentiment-negative[eng](x)\})} & \texttt{map\{$\lambda$x. random-seed1[eng](x)\}(filter\{$\lambda$x. is-sentiment-negative[eng](x)\})} & \texttt{map\{$\lambda$x. synonyms[eng](x)\}(filter\{$\lambda$x. is-sentiment-negative[eng](x)\})} \\
    \texttt{map\{$\lambda$x. entailments[eng](x)\}(filter\{$\lambda$x. is-sentiment-neutral[eng](x)\})} & \texttt{map\{$\lambda$x. random-seed1[eng](x)\}(filter\{$\lambda$x. is-sentiment-neutral[eng](x)\})} & \texttt{map\{$\lambda$x. synonyms[eng](x)\}(filter\{$\lambda$x. is-sentiment-neutral[eng](x)\})} \\
    \texttt{map\{$\lambda$x. entailments[eng](x)\}(filter\{$\lambda$x. is-sentiment-positive[eng](x)\})} & \texttt{map\{$\lambda$x. random-seed1[eng](x)\}(filter\{$\lambda$x. is-sentiment-positive[eng](x)\})} & \texttt{map\{$\lambda$x. synonyms[eng](x)\}(filter\{$\lambda$x. is-sentiment-positive[eng](x)\})} \\
    \texttt{map\{$\lambda$x. hyponyms[eng](x)\}(filter\{$\lambda$x. is-POS-adjective[eng](x)\})} & \texttt{map\{$\lambda$x. random-seed2[eng](x)\}(filter\{$\lambda$x. is-POS-adjective[eng](x)\})} & \texttt{map\{$\lambda$x. translate[eng->spa](x)\}(filter\{$\lambda$x. is-POS-adjective[eng](x)\})} \\
    \texttt{map\{$\lambda$x. hyponyms[eng](x)\}(filter\{$\lambda$x. is-POS-adverb[eng](x)\})} & \texttt{map\{$\lambda$x. random-seed2[eng](x)\}(filter\{$\lambda$x. is-POS-adverb[eng](x)\})} & \texttt{map\{$\lambda$x. translate[eng->spa](x)\}(filter\{$\lambda$x. is-POS-adverb[eng](x)\})} \\
    \texttt{map\{$\lambda$x. hyponyms[eng](x)\}(filter\{$\lambda$x. is-POS-noun[eng](x)\})} & \texttt{map\{$\lambda$x. random-seed2[eng](x)\}(filter\{$\lambda$x. is-POS-noun[eng](x)\})} & \texttt{map\{$\lambda$x. translate[eng->spa](x)\}(filter\{$\lambda$x. is-POS-noun[eng](x)\})} \\
    \texttt{map\{$\lambda$x. hyponyms[eng](x)\}(filter\{$\lambda$x. is-POS-verb[eng](x)\})} & \texttt{map\{$\lambda$x. random-seed2[eng](x)\}(filter\{$\lambda$x. is-POS-verb[eng](x)\})} & \texttt{map\{$\lambda$x. translate[eng->spa](x)\}(filter\{$\lambda$x. is-POS-verb[eng](x)\})} \\
    \texttt{map\{$\lambda$x. hyponyms[eng](x)\}(filter\{$\lambda$x. is-sentiment-negative[eng](x)\})} & \texttt{map\{$\lambda$x. random-seed2[eng](x)\}(filter\{$\lambda$x. is-sentiment-negative[eng](x)\})} & \texttt{map\{$\lambda$x. translate[eng->spa](x)\}(filter\{$\lambda$x. is-sentiment-negative[eng](x)\})} \\
    \texttt{map\{$\lambda$x. hyponyms[eng](x)\}(filter\{$\lambda$x. is-sentiment-neutral[eng](x)\})} & \texttt{map\{$\lambda$x. random-seed2[eng](x)\}(filter\{$\lambda$x. is-sentiment-neutral[eng](x)\})} & \texttt{map\{$\lambda$x. translate[eng->spa](x)\}(filter\{$\lambda$x. is-sentiment-neutral[eng](x)\})} \\
    \texttt{map\{$\lambda$x. hyponyms[eng](x)\}(filter\{$\lambda$x. is-sentiment-positive[eng](x)\})} & \texttt{map\{$\lambda$x. random-seed2[eng](x)\}(filter\{$\lambda$x. is-sentiment-positive[eng](x)\})} & \texttt{map\{$\lambda$x. translate[eng->spa](x)\}(filter\{$\lambda$x. is-sentiment-positive[eng](x)\})} \\
    \bottomrule
    \end{tabularx}
    \caption{Full list of sequential compositional tasks in \ourdataset, organized by composition type. \textit{(Continued from previous page.)}}
    \label{tab:all_tasks_seq_comp_2}
\end{table}
\end{landscape}

\end{document}